\documentclass{article}

\PassOptionsToPackage{numbers, compress}{natbib}
 \usepackage[preprint, dandb]{neurips_2025}

\usepackage[utf8]{inputenc} 
\usepackage[T1]{fontenc}    
\usepackage{hyperref}       
\usepackage{url}            
\usepackage{booktabs}       
\usepackage{amsfonts}       
\usepackage{amsmath}       
\usepackage{nicefrac}       
\usepackage{microtype}      
\usepackage{xcolor}         
\usepackage{wrapfig}        
\usepackage{bbold}          
\usepackage{caption}
\usepackage{array}
\usepackage{tabularx}
\usepackage{makecell}
\usepackage{wrapfig}
\usepackage{float}
\usepackage{footnotehyper}

\newcolumntype{Y}{>{\centering\arraybackslash}X}

\usepackage{pifont}
\usepackage{xcolor}
\newcommand{\cmark}{\ding{51}}
\newcommand{\xmark}{\ding{55}}

\usepackage{enumitem}
\makeatletter
\renewcommand{\paragraph}{%
  \@startsection{paragraph}{4}{\z@}%
                {1.5ex \@plus 0.5ex \@minus 0.2ex}%
                {-1em}%
                {\normalsize\bf}%
}
\makeatother

\title{MassSpecGym in the Wild:\\ Uncovering and Correcting Evaluation Pitfalls in AI-Driven Molecule Discovery}

\usepackage[export]{adjustbox}
\usepackage{graphicx}
\usepackage[most]{tcolorbox}
\usepackage{fancyvrb}
\usepackage{fvextra}

\tcbset{
  mytitle/.style={
    fonttitle=\bfseries\small,
    coltitle=black,
    colbacktitle=gray!10,
    sharp corners,
    boxrule=0.5pt
  },
  myprompt/.style={
    colback=white,
    colframe=black!70,
    enhanced,
    breakable
  },
  myyaml/.style={
    colback=gray!5!white,
    colframe=gray!50,
    boxrule=0.4pt
  },
}

\author{%
  \normalfont
  \makebox[0pt][c]{%
  \begin{minipage}{1.04\textwidth}
  \centering
  \small
  \textbf{Hongxuan Liu}\textsuperscript{1,*} \quad
  \textbf{Roman Bushuiev}\textsuperscript{2,3,*} \quad
  \textbf{Ivy Lightheart}\textsuperscript{4,*} \quad
  \textbf{Mrunali Manjrekar}\textsuperscript{1} \quad
  \textbf{Anton Bushuiev}\textsuperscript{2,3}\\[1pt]
  \textbf{Magdalena Lederbauer}\textsuperscript{1} \quad
  \textbf{Filip Jozefov}\textsuperscript{2} \quad
  \textbf{Yinkai Wang}\textsuperscript{5} \quad
  \textbf{Soha Hassoun}\textsuperscript{5} \quad
  \textbf{Josef Sivic}\textsuperscript{3}\\[1pt]
  \textbf{James Taylor}\textsuperscript{4} \quad
  \textbf{Runzhong Wang}\textsuperscript{1} \quad
  \textbf{David Healey}\textsuperscript{4,\textdagger} \quad
  \textbf{Tom\'a\v{s} Pluskal}\textsuperscript{2,\textdagger} \quad
  \textbf{Connor W. Coley}\textsuperscript{1,\textdagger}\\[6pt]
  \footnotesize
  \textsuperscript{1}Massachusetts Institute of Technology \quad
  \textsuperscript{2}Institute of Organic Chemistry and Biochemistry of the Czech\\[0pt]
  Academy of Sciences \quad
  \textsuperscript{3}Czech Institute of Informatics, Robotics and Cybernetics, Czech Technical\\[0pt]
  University in Prague \quad
  \textsuperscript{4}Enveda Biosciences \quad
  \textsuperscript{5}Tufts University\\[5pt]
  \textit{Correspondence to:} \texttt{david.healey@enveda.com}; \texttt{tomas.pluskal@uochb.cas.cz}; 
  \texttt{ccoley@mit.edu}\\[3pt]
  \textsuperscript{*}\,Equal contribution. \quad
  \textsuperscript{\textdagger}\,Corresponding authors.
  \end{minipage}
  }%
}

\begin{document}

\maketitle

\begin{abstract}
Reliable benchmarking is critical for developing machine learning models for tandem mass spectrometry (MS/MS)-based molecule discovery. Subtle issues in experimental design and model evaluation procedures can degrade the trustworthiness of such benchmarks and lead to erroneous conclusions. We conduct a thorough review of model evaluation issues in the recent MS/MS machine learning literature, using the standard MassSpecGym benchmark suite as a case study to illustrate the impact of these issues. We find evaluation issues in at least 18 of 26 papers reporting MassSpecGym benchmark results in the first year of its adoption. We isolate three classes of failures: (i)~data leakage, (ii)~shortcut learning, and (iii)~implementation bugs and metric divergence. Through extensive experimentation and code replication, we quantify the impact of these issues and show how they corrupt the evaluation standards MassSpecGym was designed to enforce. We distill our findings into recommendations generalizable to MS/MS challenges, benchmarks, and custom evaluation setups. We also release MassSpecGym~v1.5, an implementation of our recommendations in the MassSpecGym benchmarking suite which addresses the failure modes identified in this audit. MassSpecGym~v1.5 is publicly available at \url{https://github.com/pluskal-lab/MassSpecGym}.
\end{abstract}


\section{Introduction}
\label{sec:intro}






\begin{wrapfigure}[13]{r}{0.5\textwidth} 
    \vspace{-35pt} 
    \centering
    \includegraphics[width=\linewidth]{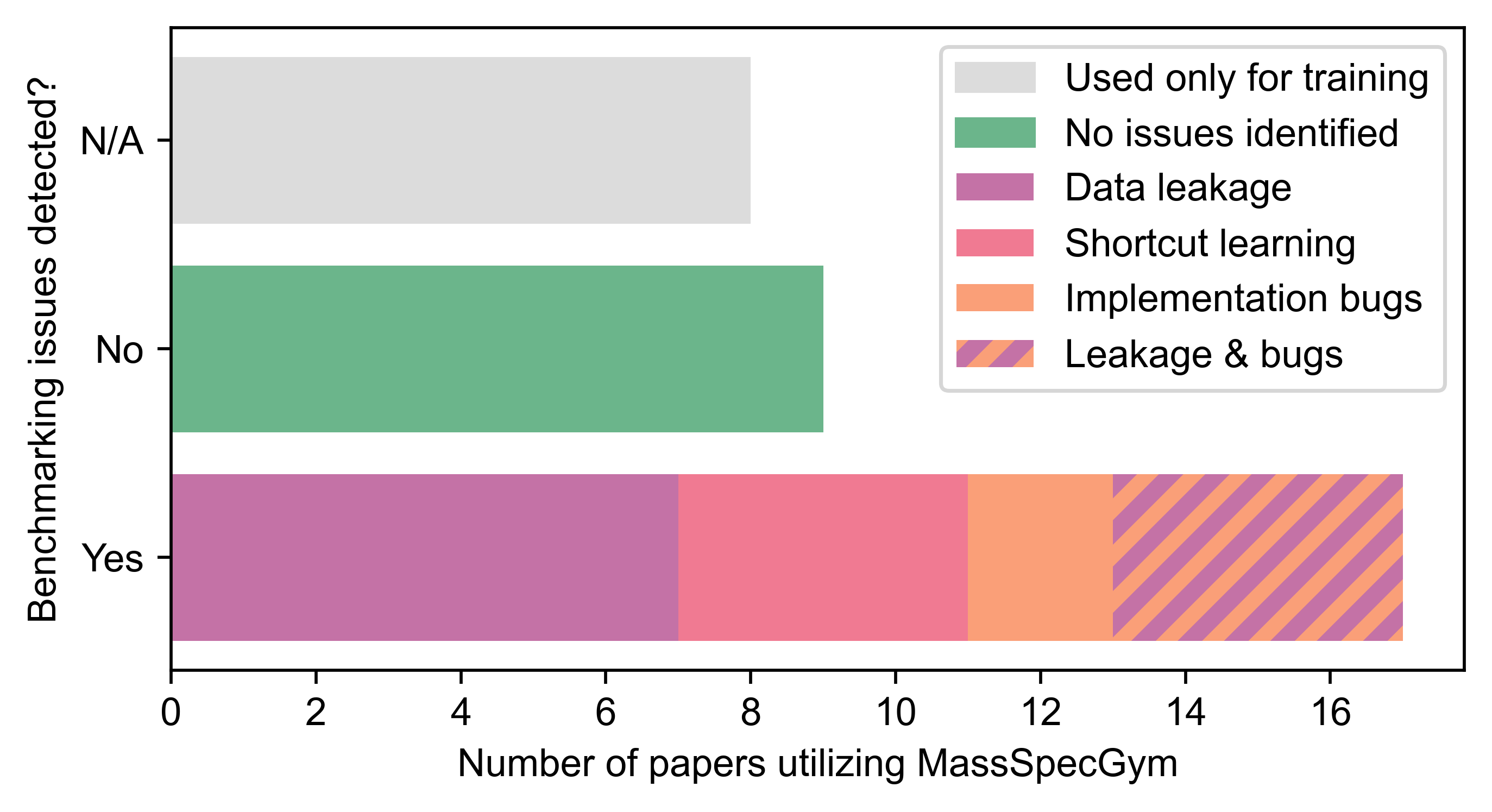}\vspace{-5pt}
    \caption{Distribution of 34 papers utilizing MassSpecGym, grouped by primary audit outcome. Benchmarking issues are pervasive throughout the first year of MassSpecGym adoption. Eight papers (gray bar) used MassSpecGym as a dataset, not as a benchmark.}
    \label{fig:monthly_audit}
\end{wrapfigure}

Machine learning is rapidly advancing the discovery and identification of small molecules given their tandem mass spectra~(MS/MS)~\citep{pollmann2026bridging}. Public benchmarks are central to this progress because they enable standardized and reproducible comparison across methods. This is especially important for molecule discovery, as confirming a novel predicted molecular structure is difficult without expensive and low-throughput experiments.~\citep{atanasov2021natural, breitmaier2002structure, stienstra2025structure}. As a result, \textit{in silico} benchmarks serve not merely as convenient proxies, but as the main operational interface through which the field measures progress.

 When benchmarks are misused, they silently reward shortcut learning, data leakage, and other causes of overfitting~\citep{liao2021are, thomas2022reliance, raji2021ai}. As benchmark-driven development accelerates, it can be unclear to what extent leaderboard gains reflect genuine progress on the underlying scientific task rather than artifacts of evaluation. The increasing automation of implementation, model reuse, and hyperparameter search further raises the standard required of benchmark design: a benchmark must be sufficiently transparent that optimization against its blind spots does not go undetected~\citep{liao2021are}. 

Over the last year, MassSpecGym~\citep{bushuiev2024massspecgym} has quickly emerged as the major benchmarking suite for machine learning in computational metabolomics. MassSpecGym formulates the problem of inferring molecular structures from tandem mass spectra as a set of standardized machine learning tasks, including: (i) the spectrum simulation challenge, in which models predict mass spectra from molecular structures; (ii) the molecule retrieval challenge, in which models identify the most compatible molecule from a candidate set given an observed mass spectrum; and (iii) the \textit{de novo} generation challenge, in which models generate molecular structures directly from a given mass spectrum.

In this paper, we conduct a systematic audit of the MassSpecGym ecosystem. Across 26 papers reporting MassSpecGym results in the first year since its introduction, we identify evaluation issues in at least 18 of them that affect the validity of their conclusions. Our goal is not to critique individual papers, but to characterize recurring failure modes and quantify their impact under corrected evaluation conditions. We organize our findings into three categories:


\textbf{1. Data leakage is more complex than a binary predicate.} Data leakage in molecular machine learning admits a gradient of stringency. Splitting on exact structure matches can yield train--test splits where the local chemical space around the test set is densely represented in training. The choice of data splitting criterion encodes a different hypothesis about what generalization in chemical space means~\citep{bushuiev2024massspecgym}. Beyond training data, we identify leakage introduced by shared components that participate in training or inference---such as formula predictors or spectral simulators---as well as data leakage from synthetic data (\S\ref{sec:contamination}).
    
\textbf{2. Retrieval tasks admit strong shortcuts unrelated to spectral reasoning.} Even with clean train--test splits, discrepancies in molecular candidate sets and composition, such as inconsistent SMILES canonicalization and  ranking biases in candidate sets, can enable shortcut learning in retrieval tasks. This can inflate hit rates (recall) to near-perfect levels without the model ever learning meaningful spectrum--molecule relationships, or in some cases without any spectral input at all (\S\ref{sec:task-validity}).
    
\textbf{3. Metric and implementation divergence propagates silently across codebases.} Errors and implementation inconsistencies in shared encoders, inference scripts, and metric implementations are often reused unchecked and adopted as convention---collectively drifting from the fair evaluation standards MassSpecGym was designed to enforce (\S\ref{sec:infrastructure}).

For each category, we present controlled experiments quantifying the impact on benchmark conclusions. We distill our findings into recommendations generalizable to mass spectrometry challenges, benchmarks, and custom evaluation setups. We consolidate the best practices into MassSpecGym~v1.5\footnote{\url{https://github.com/pluskal-lab/MassSpecGym}}, addressing 
the failure modes identified in this audit and providing canonical implementations of state-of-the-art models, 
standardized implementation of evaluation metrics, and an agentic workflow for automated auditing and continuous leaderboard maintenance (Table~\ref{tab:retrieval_leaderboard_full}) for new methods.





\section{Related Work}
\label{sec:related}
\textbf{Benchmark-driven progress and evaluation pitfalls.}
Benchmark overfitting is a common problem in machine learning, related to Goodhart’s law: optimization improves the measured objective without improving the intended one~\citep{goodhart1984problems}. This has been demonstrated through a broad literature exploring the misspecification and misuse of ML benchmarks~\citep{liao2021are, thomas2022reliance, raji2021ai, macdiarmid2025natural, recht2019do}. In particular, common evaluation pitfalls have been characterized in genomics~\citep{whalen2021navigating}, structural biology~\citep{skrinjar2025have, bushuiev2024revealing}, and small-molecule machine learning~\citep{ozcelik2025how, kretschmer2025coverage}.

\textbf{Benchmarking in MS/MS-based metabolomics.}
A number of benchmark frameworks have been proposed over the years for the evaluation of MS/MS-based machine learning models. The Critical Assessment of Small Molecule Identification (CASMI) competition has run six times since 2012~\citep{schymanski2017critical}, but its irregular schedule limits continuous method development. \citet{goldman2023annotating} created the MIST-CANOPUS benchmarking set based on GNPS spectra~\citep{wang2016sharing}, later formalizing it as NPLIB1~\citep{goldman2023prefixtree}. MassSpecGym~\citep{bushuiev2024massspecgym} advanced this direction with comprehensive public data coverage, fixed leakage-free splits, clearly defined tasks, and standardized metrics. 

Several works in recent years have performed assessments of the state of MS/MS machine learning. These include efforts to summarize recent developments~\citep{beck2024recent}, unify baselines~\citep{che2026comparative}, compare datasets and metrics~\citep{schneider2026de}, and analyze task formulation~\citep{dewaele2026small}. Other works have demonstrated the impact of problematic meta-scores~\citep{hoffmann2023mad}, made recommendations for model evaluation best practices~\citep{dejonge2022good}, and introduced new evaluation frameworks~\citep{strobel2025evaluation, zhong2026flexms}. 
While these methods focus on individual model development and benchmarking concerns, we take a broader lens: thoroughly evaluating the MassSpecGym ecosystem, classifying error modes, and experimentally demonstrating problems. 
We identify recurring evaluation problems in papers utilizing MassSpecGym and release a more robust MassSpecGym~v1.5 with concrete benchmarking recommendations. 

\section{Study Scope and Audit Protocol}
\label{sec:scope}

Our audit covers the first year of MassSpecGym adoption, targeting papers that report results on the MassSpecGym benchmark before May 1, 2026. For each paper, we inspect the reported task setting, data provenance, use of external components, metric definitions, and any publicly available implementation details or checkpoints. 

This paper combines three complementary forms of evidence. First, we conduct a structured literature and code audit to identify recurring evaluation assumptions and implementation patterns across the ecosystem. Second, we perform controlled reproductions and ablations to quantify the effect of design choices such as data splitting criteria, candidate canonicalization, and metric definitions. Third, we construct targeted stress tests to probe whether published task formulations are susceptible to adversarial or unintentional learning shortcuts. 
We focus on recurring failure modes that are both scientifically consequential and practically auditable from this first year of benchmark adoption.

\section{Data Leakage from Pretraining or External Model Components} 
\label{sec:contamination}

\emph{Data leakage} occurs when information about 
test examples enters training and inflates evaluation metrics beyond what true generalization would produce. This is problematic because performance gains are driven by memorization rather than generalizable chemical reasoning. MassSpecGym mitigates data leakage by introducing a Maximum Common Edge Subgraph (MCES)~\citep{kretschmer2025coverage} distance-based train--validation--test split that ensures molecules across data partitions are not similar more than an MCES edit distance of 10. Moreover, it introduces two explicit tracks of competition, one where formula is assumed to be known (the formula-based \textit{bonus} challenge) and the second where it is not known (the mass-based challenge). 

Data leakage can still occur when models use external pretraining data beyond MassSpecGym. Many recent \textit{de novo} generative models follow an encoder-decoder procedure: an encoder model predicts a molecular fingerprint given an MS/MS spectrum, and a decoder model generates a molecule given the predicted fingerprint. Here, we demonstrate a common pattern of MassSpecGym adoption leading to limited model generalization: using decoders~(\S\ref{sec:contamination_spectrum}) or encoders~(\S\ref{sec:transitive_contamination}) pretrained on data overlapping with the MassSpecGym test set, or on synthetic data from simulators that violate the test structure exclusion. Additionally, we show how models that use chemical-formula-based encoders leak ground-truth formulas into the mass-based challenges when trained improperly~(\S\ref{sec:formula_leakage}).
\subsection{Limited Generalization in \textit{De Novo} Models Pre-trained to Decode Fingerprints}
\label{sec:contamination_spectrum}
Many recent \textit{de novo} generation methods adopt a pretrained MIST~\citep{goldman2023annotating} spectrum-to-fingerprint encoder and focus their innovation on the fingerprint-to-molecule decoder. These decoders are typically trained on large unlabeled molecular datasets with test compounds excluded only by exact 2D InChIKey matches---a much weaker criterion than the MCES $\geq$ 10 separation that defines generalization in MassSpecGym. We hypothesize that reported gains are thus largely driven by memorization of fingerprint--molecule pairs structurally close to the test set.


\begin{figure}[t!]
\centering
\includegraphics[width=1\linewidth]{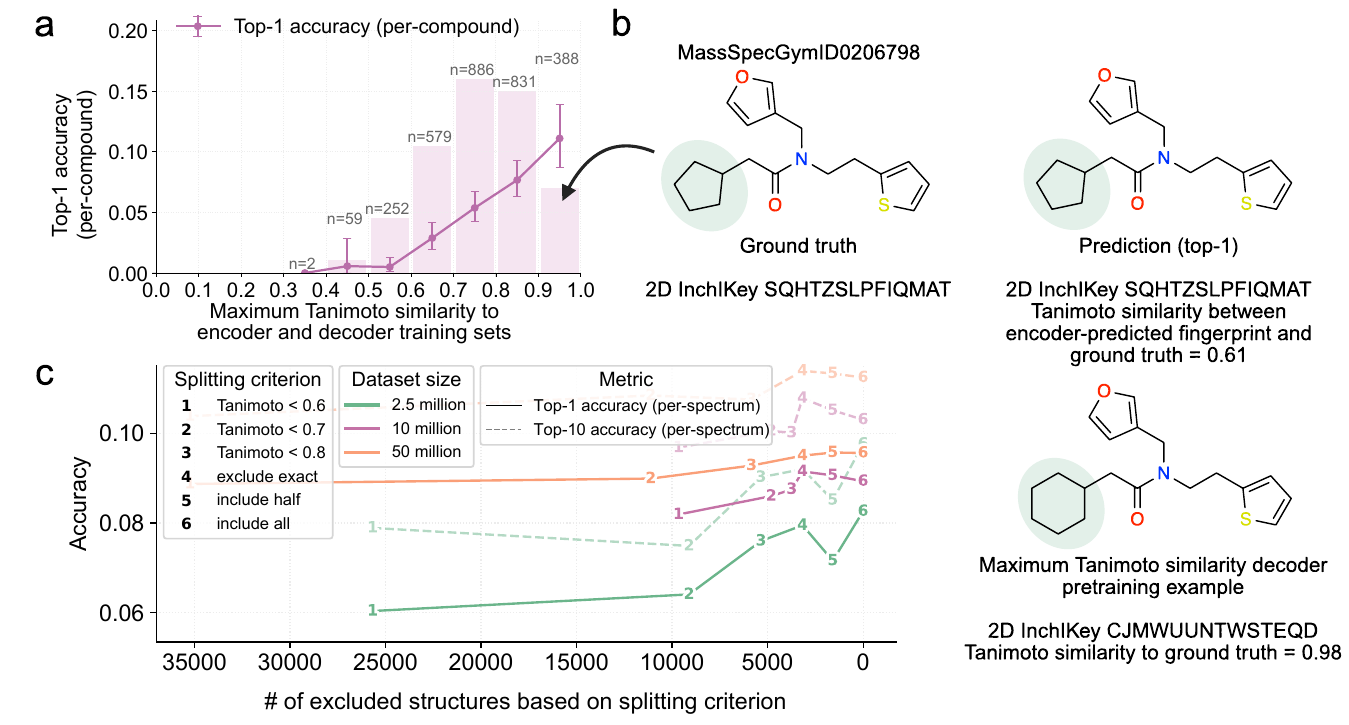}
\caption{Decoder memorization of near-identical MassSpecGym test compounds. Tanimoto similarities were computed on 2048-bit ECFP4 Morgan fingerprints (radius 2). Further experimental setup details can be found in Appendix~\ref{sec:appendix_contamination_details}. (\textbf{a}) \textit{De novo} generation accuracy correlates with similarity to the encoder and decoder training sets. Per-compound top-1 accuracy with 95\% bias-corrected and accelerated (BCa) bootstrap confidence intervals (where estimable) is shown across bins of maximum Tanimoto similarity to the union of these sets; $n$ is the per-bin count. This decoder was trained on the 50M dataset in the ``exclude exact'' setup. 
(\textbf{b}) Example of near-memorization. The ground truth (left) is reproduced perfectly by the model (top right) despite a poor encoder fingerprint prediction (Tanimoto 0.61). This is achieved by slightly modifying the closest pretraining molecule (bottom right; Tanimoto 0.98), which differs from the target at a single fingerprint bit that the encoder predicts correctly. (\textbf{c}) Benchmark sensitivity to data splitting criteria. Top-1 and top-10 accuracies for MolForge decoders trained on datasets of 2.5M, 10M, and 50M molecules across six splitting criteria.
}
\vspace{-10pt}
\label{fig:decoder_memorization}
\end{figure}


\textbf{Experiments.}
To test this hypothesis, we train a series of molecular decoders~\citep{zhao2026ms} on 2.5M molecules sampled from a curated collection of specialized databases~(Appendix~\ref{sec:appendix_contamination_details}), and on datasets of 10M and 50M molecules sampled from a combined pool of approximately 1B molecules from ZINC20~\citep{irwin2020zinc20} and UniChem~\citep{chambers2013unichem}. For each dataset, we train on six different filtering variants using progressively stricter dataset splitting criteria based on exact match exclusion and Tanimoto similarity. Each trained decoder is paired with a MIST encoder and evaluated on the formula-based MassSpecGym \textit{de novo} generation benchmark. Full condition definitions, data composition, and numerical results (including fingerprint specifications used in Tanimoto similarity calculations) are in Appendix~\ref{sec:appendix_contamination_details}.

Figure~\ref{fig:decoder_memorization} demonstrates that benchmark performance is heavily influenced by having seen close test structure analogs (for example, Tanimoto $\geq 0.7$) in the encoder or decoder training sets. In Figure~\ref{fig:decoder_memorization}a--b, we see that accuracy is substantially higher when close structure analogs to test compounds are seen during training, indicating limited generalization. This is further confirmed in Figure~\ref{fig:decoder_memorization}c, where the decoders trained on datasets of 2.5M, 10M, and 50M molecules show varying sensitivity to the filtering criterion. The 2.5M model shows additional sensitivity to training on close structure analogs to test compounds, motivating the following recommendation.

\textbf{Recommendation.}
\textit{When using external pretraining data, we recommend 
excluding exact test matches (based on the 2D InChIKey) from all train splits. If making claims about generalization, any close test structure analogs (for example, Tanimoto similarity\footnote{Using the fingerprint details mentioned in Appendix~\ref{sec:appendix_metrics}} $\geq 0.7$ or MCES $< 10$) should be excluded as well. We also recommend reporting performance stratified by the same metric between the test set and any structure in the encoder or decoder training sets (as in Figure~\ref{fig:decoder_memorization}a). To facilitate model development, MassSpecGym~v1.0 provided an official pretraining dataset of 4M curated molecules from~\citep{duhrkop2021systematic}, each at an MCES $>2$ from every molecule in the MassSpecGym validation or testing set\footnote{\url{https://huggingface.co/datasets/roman-bushuiev/MassSpecGym/blob/main/data/molecules/MassSpecGym_molecules_MCES2_disjoint_with_test_fold_4M.tsv}}. Here, we include two datasets of 2.5M and 50M molecules as official MassSpecGym~v1.5 pretraining datasets, each filtered to exclude structural analogs (Tanimoto similarity $\geq 0.7$) of any MassSpecGym test molecule\footnote{\url{https://huggingface.co/datasets/roman-bushuiev/MassSpecGym/tree/main/data/molecules}}. The official datasets also include the 1B pretraining set used by FRIGID, which excludes exact 2D InChIKey matches to MassSpecGym/NPLIB1 validation and test molecules\footnote{\label{fn:pretraining-1b}\url{https://huggingface.co/datasets/roman-bushuiev/MassSpecGym/blob/main/data/molecules/safe-gpt-v2-train-msg-canopus-val-test-inchikey14-disjoint.txt}}.}

\subsection{Spectrum-to-Fingerprint Encoders Pretrained on External Data Memorize Test Examples}
\label{sec:transitive_contamination}

Spectrum-to-fingerprint models~\citep{goldman2023annotating, stravs2022msnovelist} play an important role in many \emph{de novo} generation approaches, and are susceptible to inflated performance due to data leakage or other errors. Spectrum encoders benefit from auxiliary components like 
formula-annotation tools, which can greatly aid formula-conditioned methods; and spectral simulators~\citep{iceberg2025, alberts2024multimodal}, which can be used for synthetic data augmentation. 
Possible issues in these auxiliary components, especially when reproduced from other sources, require closer scrutiny. 
For example, we identify a commonly reproduced implementation error in the MIST encoder and discuss its impact in \S\ref{sec:mist_batching}.


We focus on the leakage risk of using synthetic spectra generated by forward models such as GLACIER~\citep{wang2026glacier} for spectrum-to-fingerprint training. The leakage risk is twofold. First, the forward model may have been trained with a data split that is not aligned with the MassSpecGym split, and thus may have learned spectrum–structure associations for molecules that appear in the MassSpecGym test set. Second, generating synthetic spectra for test molecules can expose the encoder to test-molecule-specific representations during training, effectively allowing it to overfit to the test distribution. Data leakage thus traverses GLACIER's generative process and spectrum-to-fingerprint encoder training before surfacing in metrics.

\begin{wraptable}[13]{r}{0.6\textwidth}
    \vspace{-12pt}
    \caption{\textbf{Leakage from Synthetic Spectrum Augmentation.} ``GLACIER Split” denotes the MassSpecGym partition used to train GLACIER, while ``Augmented Data Source” denotes the molecules that GLACIER generates synthetic spectra for. Setting~5 reports MIST performance without synthetic data augmentation. Retrieval uses the formula-based candidate set (i.e., w/ \textit{bonus}). Inflated results are marked in \textcolor{gray}{gray}.}
    \centering
    \vspace{-5pt}
    \resizebox{\linewidth}{!}{
    \begin{tabular}{ccccc}
    \toprule
    \# &
    \makecell{GLACIER\\Split} &
    \makecell{Augmented Data\\Source} &
    \makecell{MIST spec2fp\\Tanimoto $\uparrow$} &
    \makecell{MIST Retrieval\\Hit rate@1 $\uparrow$} \\
    \midrule
    \color{gray}{1} & \color{gray}{MSG Train + Val. + Test}  & \color{gray}{MSG-4M + MSG Test}  & \color{gray}{0.475} & \color{gray}{47.01\%} \\
        \color{gray}{2} & \color{gray}{MSG Train + Val. + Test}  & \color{gray}{MSG-4M}  & \color{gray}{0.449} & \color{gray}{43.42\%} \\
    \color{gray}{3} & \color{gray}{MSG Train} & \color{gray}{MSG-4M + MSG Test}  & \color{gray}{0.447} & \color{gray}{43.16\%} \\
    4 & MSG Train & MSG-4M & 0.433 & 40.31\% \\
    \midrule
    5 & \multicolumn{2}{c}{\makecell{MIST (no augmentation)}} & 0.262 & 9.57\% \\
    \bottomrule
    \end{tabular}
    }
    \vspace{-8pt}
    \label{tab:transitive_contamination}
\end{wraptable}
\textbf{Experiments.}
We train two GLACIER models: one on the MassSpecGym training set only, and the other on the training, validation, and test sets combined. Each GLACIER model is then used to generate synthetic spectra at five collision energies for all 4 million molecules in the MassSpecGym~v1.0 pretraining dataset, 

optionally spiked with MassSpecGym test-set structures, yielding four synthetic datasets~(Table~\ref{tab:transitive_contamination}). We train four MIST encoders on the MassSpecGym training set mixed with the respective augmented corpora and evaluate them on the MassSpecGym test set, measuring spectrum-to-fingerprint Tanimoto similarity and formula-based retrieval hit rate@1. Full setup and condition definitions are in Appendix~\ref{sec:appendix_shared_components}.

Among all settings that use synthetic data augmentation, only Setting~4---where a train-split GLACIER generates augmentation for non-test-set structures with MassSpecGym test-set molecules excluded---is leakage-free. According to Table~\ref{tab:transitive_contamination}, the other three settings exhibit varying degrees of performance inflation. In particular, Setting~1, where GLACIER is trained on all three data splits and generates test-split augmentation, substantially exceeds both the leakage-free synthetically augmented MIST performance (Setting~4) and the original MassSpecGym-reported MIST performance (Setting~5). Leakage from GLACIER's inference-time augmentation targets and training-time leakage appear to have comparable effects in this experiment, as Settings~2 and~3 exhibit similar performance inflation relative to the leakage-free Setting~4.

A related issue arises when augmentation molecules are selected using test-set information, even if exact test structures are excluded. The MIST encoder used in the earlier FRIGID evaluation was trained with ICEBERG-generated spectra for PubChem molecules selected to closely resemble the structures of MassSpecGym test molecules. Although exact test structures were excluded by 2D InChIKey, this selection still used test-set structural information \citep[Appendix B.2.3]{bohde2026frigid_v3}. FRIGID arXiv v2 reported 18.29\% top-1 accuracy \citep{bohde2026frigid_v2}; the revised evaluation with MIST \textit{4M-aug} reports 9.05\% \citep{bohde2026frigid_v3}.

\textbf{Recommendation.} \textit{We encourage developers to leverage simulated data to further scale training} only \textit{using simulators that respect the original training data split of the benchmark. Furthermore, synthetic augmentation should exclude test molecules and should not use test-set structural information to select the augmentation pool. In \S\ref{sec:mist_batching}, we release the MIST checkpoints used in Settings~4 and~5, which are trained with compliant synthetic data and without synthetic augmentation, respectively. Both checkpoints address the other issues detailed in that section, and we recommend them as the official off-the-shelf MIST checkpoints.
}

\subsection{Implicit Formula Leakage in the Mass-Based Retrieval Challenge}
\label{sec:formula_leakage}


Beyond trained spectral simulators for data augmentation, formula annotators can also be critical auxiliary components of elucidation pipelines, as formula knowledge greatly constrains the search space. While often treated as solved, formula prediction remains challenging in some cases such as halogenated molecules or molecules with masses above 600 Da. MassSpecGym differentiates along this axis: the formula-based evaluations provide ground-truth formula as input, and the mass-based evaluations withhold this information. 
In the retrieval challenge, this additionally imposes a constraint on candidate sets where the mass-based track includes candidates consistent with the observed precursor $m/z$, 
 while the formula-based track only includes formula-consistent isomers. 

Some recent methods assume known chemical formulas yet report evaluations on the mass-based tracks, leaking ground-truth formula information. Moreover, trained formula annotators that do not respect the train--test split can leak ground-truth formula information into mass-based evaluations, mirroring the transitive leakage observed with spectral simulation (\S\ref{sec:transitive_contamination}). We evaluate two possible routes of leakage through a simple filtration experiment and by training MIST-CF~\citep{Goldman2024mistcf}, a formula candidate ranker, under varying amounts of test spectrum leakage. 


\textbf{Experiments.}
We first show a lower bound on the impact of formula leakage in the mass-based challenge by pruning candidates to those whose chemical formula matches the ground truth and then selecting one at random. This modification alone improves the random baseline's hit rate@1 from 0.43\% to 12.98\% (30×~higher; Table~\ref{tab:retrieval_leaderboard_mass_challenge_w_formula}).

To demonstrate how a contaminated auxiliary formula annotator can influence downstream elucidation, 
we next train MIST-CF on the standard MassSpecGym split, 
a test-only contaminated split, and a train-and-test contaminated split. 
The formula accuracies of these models are recorded in Table~\ref{tab:mist-cf-accuracy}. Given a set of candidate formulas (we use the SIRIUS mass decomposition algorithm~\citep{Duhrkop2019sirius4}), MIST-CF ranks the candidate formulas by their agreement with the observed spectrum. We take the top-1 formula prediction on the MassSpecGym test spectra made by the three models to feed to MIST \textit{4M-aug} (which requires a molecular formula) to perform retrieval. 

Here, the predicted formulas are \textit{not} used to filter structure candidates. Full experiment details are in Appendix~\ref{app:mist-cf}.   

\begin{table*}[htbp!]
\caption{\textbf{Examination of Formula-Prediction Leakage Impact on MIST's Performance on the MassSpecGym Mass-Based Retrieval Challenge.} The first two rows indicate how ground-truth formula filtration (mirroring the distinct formula-based challenge) can boost random retrieval. The remaining rows highlight retrieval performance of MIST, which relies on chemical formula, when used with upstream MIST-CF-inferred formulas. Inflated results are marked in \textcolor{gray}{gray}. The values in parentheses indicate 99.9\% BCa bootstrap confidence intervals (20,000 resamples).}
\vspace{-0.15in}
\label{tab:retrieval_leaderboard_mass_challenge_w_formula}
\begin{center}
\renewcommand{\arraystretch}{1.15}
\small
\resizebox{\linewidth}{!}{%
\begin{tabular}{lllll}
\toprule
Method & Hit rate@1 $\uparrow$ & Hit rate@5 $\uparrow$ & Hit rate@20 $\uparrow$ & MCES@1 $\downarrow$ \\
\midrule
Random & 0.43 (0.29--0.61)\% & 2.01 (1.68--2.38)\% & 7.99 (7.34--8.74)\% & 30.91 (30.50--31.35) \\
\color{gray} Random w/ ground-truth formula & \color{gray} 12.98 (12.15--13.83)\% & \color{gray} 39.09 (37.94--40.31)\% &
\color{gray} 71.33 (70.21--72.44)\% & \color{gray} 12.46 (12.28--12.65) \\
\midrule
MIST (MIST-CF top-1 formula) & 32.14 (31.04--33.32)\% & 45.32 (44.08--46.51)\% & 60.92 (59.69--62.18)\% & 15.64
(15.27--16.02) \\
\color{gray} MIST (MIST-CF test-only top-1 formula) & \color{gray} 33.55 (32.40--34.72)\% & \color{gray} 47.85
(46.65--49.15)\% & \color{gray} 62.12 (60.89--63.41)\% & \color{gray} 14.75 (14.39--15.12) \\
\color{gray} MIST (MIST-CF train-and-test top-1 formula) & \color{gray} 36.49 (35.25--37.67)\% & \color{gray} 50.60
(49.39--51.83)\% & \color{gray} 65.60 (64.45--66.77)\% & \color{gray} 13.66 (13.32--14.01) \\
\color{gray} MIST (ground-truth formula) & \color{gray} 52.84 (51.53--54.19)\% & \color{gray} 69.86 (68.69--71.01)\% &
\color{gray} 84.17 (83.26--85.08)\% & \color{gray} 7.28 (7.03--7.54) \\
\bottomrule
\end{tabular}
}
\end{center}
\end{table*}
\vspace{-0.13in}

Here, direct or implicit leakage of ground-truth formulas can 
confer unfair benefits on downstream models that compete in the mass-based track. Using top-1 formula predictions from MIST-CF trained on both the training and test sets increases MIST's hit rate@1 from 32.14\% to 36.49\%, a gain of 4.35 percentage points (13.5\% relative improvement) over the data-safe setup. As an upper bound, access to ground-truth formulas increases the accuracy of MIST retrieval by 64.4\% (from 32.14\% to 52.84\%) compared to using the data-safe MIST-CF trained here. Though these experiments only highlight evaluations on the retrieval challenge, the same transitive leakage can also affect the \textit{de novo} challenge. As part of MassSpecGym~v1.5, we provide the data-safe version of MIST-CF 
 for use with formula-aware architectures to facilitate comparison to formula-free models. 
 

\textbf{Recommendation.} \textit{
If ground-truth formulas of }test\textit{ spectra are made available at inference, results should be reported under the formula-based track. MassSpecGym~v1.5 provides a data-safe MIST-CF formula annotator for formula-based architectures to compete in the mass-based track, and welcomes development of improved formula annotation models and strategies. }

\section{Shortcut Learning by Exploiting Retrieval Candidate Artifacts}
\label{sec:task-validity}

The MassSpecGym retrieval task~\citep{bushuiev2024massspecgym} requires a model to identify the correct ground-truth molecular structure from a candidate set of decoys, given an input MS/MS spectrum. We identify two distinct candidate-set artifacts that allow models to learn shortcuts and bypass spectral reasoning: SMILES canonicalization inconsistency~(\S\ref{sec:shortcut_canonicalization}) and ranking biases~(\S\ref{sec:shortcut_pubchem}).

\subsection{Shortcut Learning Based on Canonicalization of SMILES Strings}
\label{sec:shortcut_canonicalization}

The ground-truth SMILES strings in MassSpecGym originate directly from disparate experimental repositories (e.g., GNPS~\citep{wang2016sharing}), which may retain repository-specific formatting conventions. In contrast, the decoy candidates are uniformly collected from PubChem but canonicalized using RDKit~\citep{greg_landrum_2026_19922430}. This discrepancy introduces a 
highly discriminative distributional signature. In effect, a model can succeed by recognizing the one SMILES string formatted differently from the rest, rather than by learning the intended relationship between a mass spectrum and a molecular structure.

\vspace{-0.05in}
\begin{table*}[htbp!]
\caption{\textbf{Updated Retrieval Performance Leaderboard on the MassSpecGym Retrieval Challenge w/ \textit{Bonus} (i.e., Chemical Formula Known).} The top section presents literature-reported and updated retrieval results. The subsequent sections isolate these artifact-exploiting methods and report their corrected or spectrum-blind performance. Problematic results are marked in \textcolor{gray}{gray}.
Models where an internal reimplementation was developed are marked with a $\dag$. Further details of the Nearest Neighbor reproduction are provided in Appendix~\ref{sec:appendix_nearest_neighbor}. Code for our reproductions is available at \url{https://github.com/harrylaucngd/msms-models}. The \textit{auxiliary data} column marks methods that rely on data beyond the MassSpecGym spectrum--molecule pairs. The values in parentheses indicate 99.9\% BCa bootstrap confidence intervals (20,000 resamples).}
\vspace{-0.1in}
\label{tab:retrieval_leaderboard_full}
\begin{center}
\renewcommand{\arraystretch}{1.15}
\small
\resizebox{\linewidth}{!}{%
\begin{tabular}{lccccc}
\toprule
Method & Hit rate@1 $\uparrow$ & Hit rate@5 $\uparrow$ & Hit rate@20 $\uparrow$ & MCES@1 $\downarrow$ & Auxiliary data \\
\midrule
\multicolumn{6}{l}{\textit{Performance in the Retrieval Task w/ Bonus (i.e., Formula Known)}} \\
\midrule
Random~\citep{bushuiev2024massspecgym} & 3.06 (2.64--3.52)\% & 11.35 (10.60--12.12)\% & 27.74 (26.52--28.84)\% & 13.87 (13.70--14.03) & \xmark \\
Nearest Neighbor$^{\dag}$~\citep{khoo2026nearest} & 13.13 (12.30--13.97)\% & 26.56 (25.49--27.69)\% & 44.30 (43.14--45.59)\% & 12.45 (12.28--12.62) & \xmark \\
DeepSets~\citep{bushuiev2024massspecgym} & 4.42 (3.92--4.97)\% & 14.46 (13.58--15.36)\% & 30.76 (29.67--31.93)\% & 15.04 (14.89--15.19) & \xmark \\
Fingerprint FFN~\citep{bushuiev2024massspecgym} & 5.09 (4.57--5.66)\% & 14.69 (13.83--15.56)\% & 31.97 (30.86--33.10)\% & 14.94 (14.79--15.09) & \xmark \\
DeepSets + Fourier features~\citep{bushuiev2024massspecgym} & 6.56 (5.95--7.16)\% & 16.46 (15.58--17.35)\% & 33.46 (32.39--34.59)\% & 14.14 (13.98--14.31) & \xmark \\
MIST vMassSpecGym~\citep{bushuiev2024massspecgym} (no augmentation) & 9.57 (8.88--10.30)\% & 22.11 (21.10--23.13)\% & 41.12 (39.98--42.34)\% & 12.75 (12.59--12.91) & \xmark \\
JESTR~\citep{kalia2025jestr} & 11.82 (11.03--12.68)\% & 33.48 (32.33--34.68)\% & 61.46 (60.21--62.63)\% & 11.71 (11.54--11.87) & \xmark \\
MVP~\citep{chen2026mvp} & 13.96 (13.10--14.82)\% & 36.88 (35.71--38.02)\% & 68.12 (67.00--69.26)\% & 10.36 (10.19--10.53) & \xmark \\
FLARE~\citep{chen2026fine} & 22.66 (21.63–23.74)\% & 50.00 (48.78–51.22)\% & 75.15 (74.10–76.22)\% & 9.00 (8.82–9.18) & \xmark \\
MIST \textit{4M-aug} & 40.31 (39.05–41.47)\% & 58.87 (57.66–60.06)\% & 73.02 (71.95–74.11)\% & 7.62 (7.44–7.82) & \cmark\hyperlink{ret:note:a}{\textsuperscript{\textbf{a}}} \\
ICEBERG\footnotemark{}~\citep{iceberg2025} (forward model retrieval) & 36.18 (34.96–37.32)\% & 60.70 (59.58–61.86)\% & 78.83 (77.74–79.78)\% & 6.61 (6.43–6.79) & \xmark \\
GLACIER\hyperlink{ret:note:forward-models}{\footnotemark[\value{footnote}]}~\citep{wang2026glacier} (forward model retrieval) & 39.47 (38.20–40.65)\% & 63.50 (62.35–64.66)\% & 80.46 (79.44–81.41)\% & 6.53 (6.34–6.72) & \xmark \\
\midrule
\multicolumn{6}{l}{\textit{Exploiting Canonicalization Artifacts (Not Valid Under Corrected Candidate Canonicalization)}} \\
\midrule
\color{gray} (1) DreaMS + ChemBERTa Alignment (inflated) & \color{gray} 82.41 (81.46--83.33)\% & \color{gray} 92.57 (91.91--93.21)\% & \color{gray} 97.41 (97.00--97.78)\% & \color{gray} 2.86 (2.70--3.03) & \color{gray} \cmark \\
(1) DreaMS + ChemBERTa Alignment (corrected) & 6.70 (6.09--7.35)\% & 19.82 (18.89--20.83)\% & 40.62 (39.44--41.79)\% & 13.13 (12.97--13.28) & \cmark \\
\color{gray} (2) Chemformer Two-Stage Retrieval (inflated) & \color{gray} 63.89 (62.71--65.03)\% & \color{gray} 70.52 (69.37--71.70)\% & \color{gray} 75.86 (74.79--76.89)\% & \color{gray} 5.39 (5.20--5.60) & \color{gray} \cmark \\
(2) Chemformer Two-Stage Retrieval (corrected) & 7.60 (6.94--8.31)\% & 19.12 (18.18--20.11)\% & 39.31 (38.16--40.53)\% & 13.23 (13.06--13.39) & \cmark \\
\color{gray} (3) ChemBERTa SMILES Classifier (inflated) & \color{gray} 99.66 (99.50--99.79)\% & \color{gray} 99.84 (99.72--99.92)\% & \color{gray} 99.99 (99.95--100.00)\% & \color{gray} 0.05 (0.03--0.07) & \color{gray} \cmark \\
(3) ChemBERTa SMILES Classifier (corrected) & 8.20 (7.53--8.89)\% & 18.39 (17.38--19.36)\% & 38.14 (36.96--39.36)\% & 13.80 (13.64--13.96) & \cmark \\
\midrule
\multicolumn{6}{l}{\textit{Exploiting Other Structure Biases (Spectrum-Blind Stress Test)}} \\
\midrule
\color{gray}\textit{(4) PubChem Default Ranking} & \color{gray}\textit{49.98 (48.72--51.20)\%} &\color{gray} \textit{54.13 (52.93--55.28)\%} &\color{gray} \textit{61.04 (59.83--62.22)\%} &\color{gray} \textit{7.43 (7.22--7.64)} & \color{gray}\cmark\hyperlink{ret:note:b}{\textsuperscript{\textbf{b}}} \\
\color{gray}\textit{(5) Ranking by Number of Chiral Atoms} & \color{gray}\textit{4.88 (4.36--5.44)\%} & \color{gray}\textit{16.85 (15.92--17.79)\%} & \color{gray}\textit{37.86 (36.72--39.09)\%} & \color{gray}\textit{14.4 (14.23--14.56)} & \color{gray}\xmark \\

\bottomrule
\end{tabular}
}
\par\vspace{3pt}
\begin{minipage}{\linewidth}\footnotesize
\textsuperscript{\normalfont\hypertarget{ret:note:a}{\textbf{a}}}~The MIST \textit{4M-aug} model incorporates 4 million molecules released by MassSpecGym~v1.0 which are filtered to have an MCES distance greater than 2 from MassSpecGym test molecules, with their synthetic spectra during training (see \S\ref{sec:transitive_contamination}).
\textsuperscript{\normalfont\hypertarget{ret:note:b}{\textbf{b}}}~PubChem default ranking uses the PubChem PUG-REST API's default return order~(see Appendix~\ref{sec:appendix_jailbreak}).
\end{minipage}
\end{center}
\vskip -0.1in
\end{table*}

\textbf{Experiments.}
To quantify the magnitude of this shortcut effect, we evaluated several recent architectures and constructed an explicit artifact-exploiting baseline (see Appendix~\ref{sec:appendix_jailbreak} for complete formulations). Table~\ref{tab:retrieval_leaderboard_full} reports the results. {(1) DreaMS + ChemBERTa Alignment}: a DreaMS spectral encoder~\citep{bushuiev2025dreams} is aligned with a frozen ChemBERTa molecular encoder~\citep{ahmad2022chemberta2} via a bidirectional InfoNCE loss; because the ground-truth SMILES strings retain non-canonical syntax while all decoys are RDKit-canonicalized, the model can achieve over 82\% hit rate@1 by matching formatting cues rather than learning true structure--spectrum correspondence. {(2) Chemformer Two-Stage Retrieval}: a frozen Chemformer encoder~\citep{irwin2022chemformer} is paired with a trained spectral encoder via dual-path InfoNCE for initial re-ranking~(Stage~1), followed by a cross-fusion decoder that autoregressively generates SMILES conditioned on the top-$K$ retrieved candidates and uses them to refine the ranking~(Stage~2); the same canonicalization discrepancy biases from Stage~1 also score true molecules higher, an effect that cascades into Stage~2 to achieve over 63\% hit rate@1. {(3) ChemBERTa SMILES Classifier}: a format classifier trained only on SMILES syntax to distinguish query molecules from their retrieval candidates achieves near-perfect hit rate@1.
\footnotetext{\hypertarget{ret:note:forward-models}{}ICEBERG and GLACIER are evaluated on the full MassSpecGym test split, roughly half of which lack collision energy annotations. We imputed these missing collision energies; details of the annotation pipeline are given in Appendix~\ref{sec:appendix_collsion_energy}.}

\textbf{Recommendation.} \textit{Uniform RDKit canonicalization should be enforced across all candidate and query SMILES before training and evaluation. 
MassSpecGym~v1.5 applies this correction by default.\footnote{\url{https://huggingface.co/datasets/roman-bushuiev/MassSpecGym}}}
\subsection{Shortcut Learning via PubChem Ranking Biases}
\label{sec:shortcut_pubchem}

The chemical space of structures with measured MS/MS spectra is inherently non-uniform and exhibits a distinct distribution from PubChem (or other potential sources of decoys). Ground-truth molecules are typically common metabolites, widely available standards, or easily synthesizable compounds. They tend to appear with higher deposition frequency, and may share structural features that distinguish them from arbitrary entries in candidate pools drawn from compound libraries such as PubChem. A model can exploit these spurious correlations to retrieve the correct candidate while ignoring the MS/MS spectrum entirely.

\textbf{Experiments.}
To empirically demonstrate shortcut learning from candidate pools, we evaluate two spectrum-free baselines (Table~\ref{tab:retrieval_leaderboard_full}; implementation details are available in Appendix~\ref{sec:appendix_jailbreak}). {(4) PubChem Default Ranking}: ranking candidates by PubChem PUG-REST API's default return order, ignoring the spectrum entirely, can exceed 49\% hit rate@1. {(5) Ranking by Number of Chiral Atoms}: a spectrum-blind baseline ranking candidate molecules by the number of chiral centers performs on par with simple learnable fingerprint prediction baselines. More generally, the distribution shift between query and candidate PubChem molecules is apparent from the performance of the corrected baseline {(3)}, which achieves 8.2\% hit rate@1 by directly learning the distributional artifacts that distinguish them.

While the canonicalization artifact can be straightforwardly fixed by enforcing uniform RDKit canonicalization across all query and candidate molecules (which we implement in MassSpecGym~v1.5), the ranking bias and spurious correlations between queries and candidates~(e.g., num. of chiral atoms) represent a more profound challenge for the community: molecules with available experimental spectra for benchmarking are not uniformly distributed across the chemical space of compound libraries such as PubChem. Instead, they disproportionately fall into sub-regions comprising molecules that are more routinely synthesized, characterized, and analyzed. We therefore treat this as an open evaluation-design question: how should retrieval benchmarks disentangle a model's true spectral reasoning capability from shortcut learning over query and candidate molecules? 

In structure-based drug design, the fact that property-matched decoys are harder to distinguish from true binders than random decoys has been discussed for decades \cite{huang2006benchmarking, mysinger2012directory}. Analogously, in their work concurrent to this article, \citet{gupta2026confronting} showed that using property-matched decoys for the MassSpecGym retrieval benchmark reduces the performance of a spectrum-blind baseline model from competitive to near random.
This motivates future benchmarking design in computational metabolomics to alleviate distribution shifts with property-matched decoy structures.

\textbf{Recommendation.} \textit{Authors of new methods should verify that their models do not rely on spurious correlations between query and candidate molecules---for example, by masking all input MS/MS spectra with a constant value and confirming that performance degrades accordingly.}

\section{Implementation Bugs and Metric Divergence Undermine Reproducibility}
\label{sec:infrastructure}

Benchmark failures are not limited to data curation and task design. A subtler class of problems emerges at evaluation time in the form of implementation bugs and configuration inconsistencies. These may only be apparent when auditing the code.

As MassSpecGym adoption has grown, its evaluation ecosystem has become increasingly modular: pretrained encoders are reused as frozen feature extractors, inference scripts are copied across codebases, and configuration choices made in one project are inherited, often without examination, by the next. We show that excessive reuse can propagate silent errors. Undocumented default settings in shared components carry over to downstream methods and inflate their reported performance. Separately, we show that implementation-level divergence in metric computation---fingerprint radius or bit length for Tanimoto similarity, stereochemistry handling in InChIKey matching, or threshold and bound settings for MCES distance---is sufficient to reorder leaderboard rankings when metric settings differ across models, even with predictions held fixed. 

\subsection{The MIST Inference Batching Pitfall}
\label{sec:mist_batching}


MIST~\citep{goldman2023annotating} has become a widely reused encoder in the MassSpecGym ecosystem, adopted as a feature extractor by several subsequent works~\citep{bohde2025diffms, neo2025molforge}. Our audit found two distinct MIST-related failure modes: a conventional padding-mask error in batched settings, and a separate checkpoint-provenance problem in one specific commonly reused MIST model.


\subsubsection{Padding Mask Bug in Batched MIST}
\label{subsec:mist_batching}

The original MIST encoder was validated in single-spectrum settings. When inputs are batched, its attention code as 
initially published in the MIST and DiffMS codebases mistakenly applied the raw boolean mask via \texttt{attn += attn\_mask}. Padding tokens therefore receive a finite $+1$ attention offset rather than being excluded from the softmax, affecting both training and inference with padded batches.

\subsubsection{Unreproducible MIST vDiffMS Checkpoint Affecting Downstream Decoder Development}
\vspace{-3pt}
\begin{table*}[htbp!]
\caption{\textbf{Comparison of Correct and Problematic MIST + MolForge Performances on the MassSpecGym \textit{de novo} Generation Challenge w/ \textit{Bonus}.} $^\dag$ indicates that the method was reimplemented. Problematic results are marked in \textcolor{gray}{gray}.
}
\vspace{-0.15in}
\label{tab:massspecgym_leaderboard}
\begin{center}
\renewcommand{\arraystretch}{1.15}
\small
\resizebox{\linewidth}{!}{%
\begin{tabular}{lccccccc}
\toprule
\multicolumn{1}{c}{} & \multicolumn{3}{c}{Top-1} & \multicolumn{3}{c}{Top-10} & \multicolumn{1}{c}{Encoder} \\
\cmidrule(lr){2-4}
\cmidrule(lr){5-7}
\cmidrule(lr){8-8}
Models and results on \textit{de novo} task & Accuracy $\uparrow$ & MCES $\downarrow$ & Tanimoto $\uparrow$ & Accuracy $\uparrow$ & MCES $\downarrow$ & Tanimoto $\uparrow$ & MIST Tanimoto $\uparrow$ \\
\midrule
MIST \textit{4M-aug} + MolForge$^\dag$ & 3.29\% & 31.77 & 0.31 & 5.65\% & 16.67 & 0.40 & 0.433 \\
\color{gray} MIST vDiffMS + MolForge$^\dag$ (batch size = 1) & \color{gray} 10.73\% & \color{gray} 22.15 & \color{gray} 0.37 & \color{gray} 14.48\% & \color{gray} 17.88 & \color{gray} 0.41 & \color{gray} 0.457 \\
\color{gray} MIST vDiffMS + MolForge$^\dag$ (batch size = 24) & \color{gray} 31.75\% & \color{gray} 12.30 & \color{gray} 0.68 & \color{gray} 40.55\% & \color{gray} 9.80 & \color{gray} 0.74 & \color{gray} 0.812 \\
\bottomrule
\end{tabular}
}
\end{center}
\vskip -0.1in
\end{table*}

The subsequent erroneous MIST vDiffMS checkpoint trained and used for inference as described in Section~\ref{subsec:mist_batching} 
 produces fingerprint predictions, at least in batched mode, that are consistent with patterns of data leakage or contamination; however, we cannot reproduce this inflation from the documented training recipe. 
Moreover, this erroneous MIST checkpoint has been reused in several \textit{de novo} generation pipelines, inflating spectrum-to-fingerprint performance far above what we obtain from a freshly trained, data-safe MIST encoder. 
As Table~\ref{tab:massspecgym_leaderboard} shows, replacing batched inference settings with single-spectrum settings sharply reduces downstream MIST + MolForge performance. Given the irreproducibility of MIST vDiffMS, we introduce and recommend the MIST \textit{4M-aug} checkpoint, trained on the MassSpecGym training set mixed with synthetic spectra generated by GLACIER for the official 4 million unlabeled molecules released with MassSpecGym~v1.0. Our aim is to prevent further propagation of unreproducible or otherwise problematic MIST checkpoints and provide a more unified and trustworthy basis for measuring the decoder-side capabilities of \textit{de novo} methods. Full training details for MIST \textit{4M-aug} are provided in Appendix~\ref{sec:mist}. Table~\ref{tab:massspecgym_leaderboard_full} provides the full updated performance leaderboard for the \textit{de novo} \textit{bonus} task.

\textbf{Recommendation.} \textit{Practice careful review when adopting existing model components or other code, especially for use in development settings they were not designed for. Unintended bugs can occur during training, end-to-end fine-tuning, and/or inference. 
MassSpecGym~v1.5 includes the corrected MIST implementation.\footnote{\url{https://github.com/coleygroup/MIST}} We recommend MIST \textit{4M-aug} as the official off-the-shelf MIST checkpoint for future benchmarking.}
\vspace{-1pt}
\subsection{Metric Implementation Divergence Hindering Reproducibility}
\vspace{-1pt}
A second reproducibility failure stems from metric implementations that vary across codebases. We identify four metrics---InChIKey hit rate, MCES distance, Tanimoto fingerprint similarity, and close/meaningful match rates---requiring implementation choices that are neither standardized across works nor consistently documented (complete formal definitions are given in Appendix~\ref{sec:appendix_metrics}).

\begin{savenotes}
\begin{table}[H]
\caption{\textbf{Updated Leaderboard on the MassSpecGym \textit{de novo} Generation Bonus Challenge.} This table covers methods first reporting MassSpecGym results before May 1, 2026, and includes revised evaluations. The best-performing model for each metric is \textbf{bold} and the second best is \underline{underlined}. $\dag$~indicates our implementations of baseline approaches without public code. Gray rows are excluded from the corrected ranking: MIST + MolForge uses the unreproducible MIST vDiffMS checkpoint (\S\ref{sec:mist_batching}), while FRIGID arXiv v2 uses test-informed synthetic augmentation (\S\ref{sec:transitive_contamination}). MCES calculations for MSNovelist, Neuraldecipher, and DiffMS were recomputed from the original DiffMS paper in~\citep{manjrekar2026generative} to correctly use a myopic threshold of 15 instead of 100. For MS-BART, we report accuracy based on 2D InChIKey matching from Appendix~C of~\citep{han2025msbart}, rather than the full InChIKey matching accuracy reported in the main text. FRIGID uses 10 rounds of ICEBERG-guided inference-time refinement. The \textit{auxiliary data} column marks methods that rely on data beyond the MassSpecGym spectrum--molecule pairs. Code for our reproductions is available at \url{https://github.com/harrylaucngd/msms-models}.}
\vspace{-0.05in}
\label{tab:massspecgym_leaderboard_full}
\begin{center}
\renewcommand{\arraystretch}{1.15}
\small
\resizebox{\linewidth}{!}{%
\begin{tabular}{lccccccc}
\toprule
& \multicolumn{3}{c}{Top-1} & \multicolumn{3}{c}{Top-10} & \\
\cmidrule(lr){2-4}
\cmidrule(lr){5-7}
Model & Accuracy $\uparrow$ & MCES $\downarrow$ & Tanimoto $\uparrow$ & Accuracy $\uparrow$ & MCES $\downarrow$ & Tanimoto $\uparrow$ & Auxiliary data \\
\midrule
SMILES Transformer~\citep{bushuiev2024massspecgym} & 0.00\% & 79.39 & 0.03 & 0.00\% & 52.13 & 0.10 & \xmark \\
SELFIES Transformer~\citep{bushuiev2024massspecgym} & 0.00\% & 38.88 & 0.08 & 0.00\% & 26.87 & 0.13 & \xmark \\
Random generation~\citep{bushuiev2024massspecgym} & 0.00\% & 21.11 & 0.08 & 0.00\% & 18.26 & 0.11 & \xmark \\
MIST + MSNovelist~\citep{stravs2022msnovelist, bohde2025diffms} & 0.00\% & 39.84 & 0.06 & 0.00\% & 18.83 & 0.15 & \cmark\hyperlink{tn:a}{\textsuperscript{\textbf{a}}} \\
Spec2Mol~\citep{Litsa2023deep, bohde2025diffms} & 0.00\% & 36.78 & 0.12 & 0.00\% & 36.02 & 0.16 & \xmark \\
MIST + Neuraldecipher~{\citep{Le2020neuraldecipher, bohde2025diffms}} & 0.00\% & 22.93 & 0.14 & 0.00\% & 21.76 & 0.16 & \cmark\hyperlink{tn:a}{\textsuperscript{\textbf{a}}} \\
MADGEN~\citep{wang2025madgen} & 1.31\% & 27.47 & 0.20 & 1.54\% & 16.84 & 0.26 & \cmark\hyperlink{tn:b}{\textsuperscript{\textbf{b}}} \\
MS-BART~\citep{han2025msbart} & 1.26\% & 16.47 & 0.23 & 1.28\% & 15.12 & 0.28 & \cmark\hyperlink{tn:c}{\textsuperscript{\textbf{c}}} \\
DiffMS~\citep{bohde2025diffms} & 2.30\% & 13.96 & 0.28 & 4.25\% & 11.68 & 0.39 & \cmark\hyperlink{tn:a}{\textsuperscript{\textbf{a}}} \\
FOAM (DiffMS + PubChem)~\citep{manjrekar2026generative} & 1.50\% & \textbf{12.21} & 0.35 & \underline{10.28\%} & \textbf{6.06} & \textbf{0.53} & \cmark\hyperlink{tn:a}{\textsuperscript{\textbf{a}}}$^{,}$\hyperlink{tn:d}{\textsuperscript{\textbf{d}}} \\
MSAnchor~\citep{Qin2026MSAnchorDN} & 2.68\% & 16.57 & 0.32 & 4.67\% & 14.12 & 0.41 & \cmark\hyperlink{tn:e}{\textsuperscript{\textbf{e}}} \\
MIST + MolForge~\citep{neo2025molforge}$^\dag$ (corrected) & 3.29\% & 31.77 & 0.31 & 5.65\% & 16.67 & 0.40 & \cmark\hyperlink{tn:a}{\textsuperscript{\textbf{a}}}$^{,}$\hyperlink{tn:f}{\textsuperscript{\textbf{f}}} \\
\color{gray} MIST + MolForge$^\dag$ (inflated, batch size = 24) & \color{gray} 31.75\% & \color{gray} 12.30 & \color{gray} 0.68 & \color{gray} 40.55\% & \color{gray} 9.80 & \color{gray} 0.74 & \color{gray} \cmark\hyperlink{tn:a}{\textsuperscript{\textbf{a}}}$^{,}$\hyperlink{tn:f}{\textsuperscript{\textbf{f}}} \\
FRIGID-base~\citep{bohde2026frigid_v3} (MIST \textit{4M-aug}) & \underline{8.21\%} & 13.16 & \underline{0.38} & 9.82\% & 11.65 & \underline{0.43} & \cmark\hyperlink{tn:f}{\textsuperscript{\textbf{f}}}$^{,}$\hyperlink{tn:g}{\textsuperscript{\textbf{g}}} \\
\makecell[l]{FRIGID~\citep{bohde2026frigid_v3} (MIST \textit{4M-aug})} & \textbf{9.05\%} & \underline{12.60} & \textbf{0.39} & \textbf{12.23\%} & \underline{11.41} & \underline{0.43} & \cmark\hyperlink{tn:f}{\textsuperscript{\textbf{f}}}$^{,}$\hyperlink{tn:g}{\textsuperscript{\textbf{g}}} \\
\color{gray} FRIGID~\citep{bohde2026frigid_v2} (arXiv v2) & \color{gray} 18.29\% & \color{gray} 13.49 & \color{gray} 0.43 &
  \color{gray} 22.00\% & \color{gray} 11.65 & \color{gray} 0.47 & \color{gray} \cmark\hyperlink{tn:f}{\textsuperscript{\textbf{f}}}$^{,}$\hyperlink{tn:g}{\textsuperscript{\textbf{g}}} \\
\bottomrule
\end{tabular}%
}
\par\vspace{3pt}
\begin{minipage}{\linewidth}\footnotesize
\textsuperscript{\normalfont\hypertarget{tn:a}{\textbf{a}}}~The MIST + MSNovelist, MIST + Neuraldecipher, DiffMS, and MIST + MolForge decoders were pretrained on 2.8 million curated molecules\footnote{\url{https://github.com/coleygroup/DiffMS/blob/master/data_processing/00_download_fp2mol_data.sh}}, 2D InChIKey-disjoint with MassSpecGym test and validation molecules~(see \cite{bohde2025diffms}). \textsuperscript{\normalfont\hypertarget{tn:b}{\textbf{b}}}~MADGEN ranks Murcko scaffolds over the 256 formula-matched candidate structures provided by the MassSpecGym retrieval challenge~(see \cite{wang2025madgen}). \textsuperscript{\normalfont\hypertarget{tn:c}{\textbf{c}}}~MS-BART was pretrained on the 4 million unlabeled molecules released with MassSpecGym~v1.0, at an MCES distance $>2$ from MassSpecGym test molecules~(see~\cite{han2025msbart}). \textsuperscript{\normalfont\hypertarget{tn:d}{\textbf{d}}}~At inference, FOAM seeds its optimization with DiffMS-based suggestions and formula-matched PubChem structures, with the latter set 2D InChIKey-disjoint with MassSpecGym test molecules~(see \cite{manjrekar2026generative}). \textsuperscript{\normalfont\hypertarget{tn:e}{\textbf{e}}}~The first-stage Transformer in MSAnchor was pretrained on spectra from SDBS~\citep{sdbs} and the multimodal spectroscopic dataset of Alberts et al.~\citep{alberts2024multimodal}, excluding molecules with MCES distance $<11$ from any test molecule~(see \cite{Qin2026MSAnchorDN}). \textsuperscript{\normalfont\hypertarget{tn:f}{\textbf{f}}}~The gray MIST + MolForge and FRIGID rows use the earlier MIST checkpoints discussed in \S\ref{sec:mist_batching} and \S\ref{sec:transitive_contamination}, respectively. The corrected MIST + MolForge and FRIGID results use MIST \textit{4M-aug}, trained on MassSpecGym training spectra and GLACIER-generated spectra for the official 4 million unlabeled molecules released with MassSpecGym~v1.0. These corrected results are evaluated on all 17{,}556 test spectra (Appendix~\ref{sec:appendix_contamination_details}). \textsuperscript{\normalfont\hypertarget{tn:g}{\textbf{g}}}~The FRIGID decoder was pretrained from the 1-billion-molecule corpus of~\cite{noutahi2023gotta}, with additional filtering to exclude structures sharing a 2D InChIKey with any MassSpecGym/NPLIB1 test or validation molecule\textsuperscript{\ref{fn:pretraining-1b}}~(see \cite{bohde2026frigid_v3}).
\end{minipage}
\end{center}
\vskip -0.1in
\end{table}
\end{savenotes}

\textbf{InChIKey}: as spectral information often cannot disambiguate stereochemistry, deduplication and accuracy evaluation should only use the 14-character, 2D connectivity layer; using the full 27-character key treats stereoisomers as distinct molecules and undercounts true overlaps. For example, full InChIKey matching lowers FRIGID's top-1/top-10 accuracy from 9.05\%/12.23\% to 4.97\%/6.80\% (Table~\ref{tab:metric_variants}). \textbf{MCES}: the distance $|E(G_1)|+|E(G_2)|-2|\text{MCES}(G_1,G_2)|$ is NP-hard and computed via heuristic solvers whose timeout and approximation parameters vary across implementations. \textbf{Tanimoto similarity}: computed on molecular fingerprints, the choice of fingerprint algorithm and parameters like radius and bit length can significantly change Tanimoto similarities.  

\textbf{Recommendation.} \textit{Pinned, standardized metric implementations should be used: 14-character connectivity-layer matching for InChIKey; 2048-bit ECFP4 without chirality for Tanimoto similarity; \texttt{myopic-mces}~1.0.0 with \texttt{threshold=15} and \texttt{always\_stronger\_bound=True} for MCES.
We further recommend reporting mean test-set metrics with 99.9\% BCa confidence intervals computed from 20,000 bootstrap resamples.
MassSpecGym~v1.5 provides reference implementations for all metrics; metric code copied from third-party repositories should be independently verified against the benchmark specification.}

\vspace{-3pt}
\section{Discussion and Release of MassSpecGym~v1.5}
\label{sec:discussion}
\vspace{-3pt}

The evaluation failures documented in the preceding sections have been highlighted in the context of MassSpecGym, but are broadly relevant to computational mass spectrometry. 
In several cases, these effects produce unexpectedly strong performance, underscoring the need for skepticism toward results that appear too good to be true. 
The failure modes we have identified have motivated specific recommendations that will promote more defensible evaluations for this rapidly growing field. 

\textbf{MassSpecGym~v1.5 as a benchmark revision.}
The revised benchmark accompanying this paper is MassSpecGym~v1.5, which operationalizes the recommendations and findings discussed above. 
\textbf{At the data level}, we perform uniform RDKit canonicalization across all candidate SMILES, eliminating the formatting shortcut. In addition to the dataset of 4M molecules from MassSpecGym~v1.0, we also release new 2.5M and 50M decoder pretraining sets that exclude close structure analogs (for example, Tanimoto similarity $\geq0.7$) to any test compound, providing a safer and more interpretable foundation for generative model training. \textbf{At the model level}, we release a zoo of models including officially retrained data-safe versions of MIST~\citep{goldman2023annotating}, MIST-CF~\citep{Goldman2024mistcf}, ICEBERG~\citep{iceberg2025}, GLACIER~\citep{wang2026glacier} and DreaMS~\citep{bushuiev2025dreams} to provide leakage-free components for formula prediction, spectral simulation, and MS/MS embeddings computation. We also clarify previously ambiguous evaluation procedures and correct identified errors, updating the leaderboards for both the retrieval and \textit{de novo} \textit{bonus} tasks (Tables~\ref{tab:retrieval_leaderboard_full} and~\ref{tab:massspecgym_leaderboard_full}). These updates standardize evaluation criteria, correct inflated retrieval results from problematic implementations, and address encoder-induced performance inflation in the \textit{de novo} task. The resulting changes in absolute \textit{de novo} scores do not substantially alter the relative ranking of the evaluated methods. \textbf{At the evaluation level}, we package the pinned metric implementations, corrected inference settings, reference results used throughout this audit, and an LLM-augmented agentic workflow for code review outlined in Appendix~\ref{sec:appendix_code-review}, so that future work can be compared under the same benchmark assumptions.  Together, these updates reduce ambiguity in how MassSpecGym results are produced, interpreted, and reproduced.

\textbf{Limitations and broader implications.}
The three-category audit framework we apply---data leakage analysis, task validity stress-testing, and reproducibility verification---is likely to transfer to related benchmark ecosystems. At the same time, our analysis has important limitations. First, the audit covers only the first year of MassSpecGym adoption to date (34 papers), and evaluation practice may change as the community matures and more model implementations are made publicly available. Second, our experiments (particularly the pretraining data leakage spectrum in \S\ref{sec:contamination}) illustrate trends in performance under example design choices, but the quantitative results and implications (e.g., recommended similarity thresholds) may shift as more models and corpora are analyzed. Third, although MassSpecGym~v1.5 corrects several major sources of ambiguity, it does not resolve every evaluative challenge exposed by this audit---most notably, the ranking bias in candidate sets (\S\ref{sec:task-validity}) remains an open design problem inherent to the retrospective nature of this, or any, benchmark.



\bibliography{bibliography}

@inproceedings{bushuiev2024massspecgym,
  author       = {Roman Bushuiev and
                  Anton Bushuiev and
                  Niek F. de Jonge and
                  Adamo Young and
                  Fleming Kretschmer and
                  Raman Samusevich and
                  Janne Heirman and
                  Fei Wang and
                  Luke Zhang and
                  Kai D{\"{u}}hrkop and
                  Marcus Ludwig and
                  Nils A. Haupt and
                  Apurva Kalia and
                  Corinna Brungs and
                  Robin Schmid and
                  Russell Greiner and
                  Bo Wang and
                  David S. Wishart and
                  Liping Liu and
                  Juho Rousu and
                  Wout Bittremieux and
                  Hannes Rost and
                  Tytus D. Mak and
                  Soha Hassoun and
                  Florian Huber and
                  Justin J. J. van der Hooft and
                  Michael A. Stravs and
                  Sebastian B{\"{o}}cker and
                  Josef Sivic and
                  Tom{\'{a}}s Pluskal},
  editor       = {Amir Globersons and
                  Lester Mackey and
                  Danielle Belgrave and
                  Angela Fan and
                  Ulrich Paquet and
                  Jakub M. Tomczak and
                  Cheng Zhang},
  title        = {{MassSpecGym}: {A} benchmark for the discovery and identification of
                  molecules},
  booktitle    = {Advances in Neural Information Processing Systems 38: Annual Conference
                  on Neural Information Processing Systems 2024, NeurIPS 2024, Vancouver,
                  BC, Canada, December 10 - 15, 2024},
  year         = {2024},
  url          = {http://papers.nips.cc/paper\_files/paper/2024/hash/c6c31413d5c53b7d1c343c1498734b0f-Abstract-Datasets\_and\_Benchmarks\_Track.html},
  bibsource    = {dblp computer science bibliography, https://dblp.org}
}

@Article{dejonge2022good,
author={de Jonge, Niek F.
and Mildau, Kevin
and Meijer, David
and Louwen, Joris J. R.
and Bueschl, Christoph
and Huber, Florian
and van der Hooft, Justin J. J.},
title={Good practices and recommendations for using and benchmarking computational metabolomics metabolite annotation tools},
journal={Metabolomics},
year={2022},
month=dec,
day={05},
volume={18},
number={12},
pages={103},
issn={1573-3890},
doi={10.1007/s11306-022-01963-y},
url={https://doi.org/10.1007/s11306-022-01963-y}
}

@Article{strobel2025evaluation,
author={Strobel, Michael
and Gil-de-la-Fuente, Alberto
and Zare Shahneh, Mohammad Reza
and Abiead, Yasin El
and Bushuiev, Roman
and Bushuiev, Anton
and Pluskal, Tom{\'a}{\v{s}}
and Wang, Mingxun},
title={An evaluation methodology for machine learning-based tandem mass spectra similarity prediction},
journal={BMC Bioinformatics},
year={2025},
month=jul,
day={11},
volume={26},
number={1},
pages={174},
issn={1471-2105},
doi={10.1186/s12859-025-06194-1},
url={https://doi.org/10.1186/s12859-025-06194-1}
}

@Article{schymanski2017critical,
author={Schymanski, Emma L.
and Ruttkies, Christoph
and Krauss, Martin
and Brouard, C{\'e}line
and Kind, Tobias
and D{\"u}hrkop, Kai
and Allen, Felicity
and Vaniya, Arpana
and Verdegem, Dries
and B{\"o}cker, Sebastian
and Rousu, Juho
and Shen, Huibin
and Tsugawa, Hiroshi
and Sajed, Tanvir
and Fiehn, Oliver
and Ghesqui{\`e}re, Bart
and Neumann, Steffen},
title={Critical Assessment of Small Molecule Identification 2016: automated methods},
journal={Journal of Cheminformatics},
year={2017},
month=mar,
day={27},
volume={9},
number={1},
pages={22},
issn={1758-2946},
doi={10.1186/s13321-017-0207-1},
url={https://doi.org/10.1186/s13321-017-0207-1}
}

@Article{goldman2023annotating,
author={Goldman, Samuel
and Wohlwend, Jeremy
and Stra{\v{z}}ar, Martin
and Haroush, Guy
and Xavier, Ramnik J.
and Coley, Connor W.},
title={Annotating metabolite mass spectra with domain-inspired chemical formula transformers},
journal={Nature Machine Intelligence},
year={2023},
month=sep,
day={01},
volume={5},
number={9},
pages={965-979},
issn={2522-5839},
doi={10.1038/s42256-023-00708-3},
url={https://doi.org/10.1038/s42256-023-00708-3}
}

@ARTICLE{chen2026fine,
  title       = "{FLARE}: Fine-grained learning for alignment of
                 spectra-molecule {REpresentation} enhances metabolite
                 annotation",
  author      = "Chen, Yan Zhou and Rushing, Blake and Hassoun, Soha",
  journal     = "bioRxiv",
  institution = "bioRxiv",
  pages       = "2026.01.27.702086",
  month       =  jan,
  year        =  2026,
  language    = "en"
}

@inproceedings{goldman2023prefixtree,
  author       = {Samuel Goldman and
                  John Bradshaw and
                  Jiayi Xin and
                  Connor W. Coley},
  editor       = {Alice Oh and
                  Tristan Naumann and
                  Amir Globerson and
                  Kate Saenko and
                  Moritz Hardt and
                  Sergey Levine},
  title        = {Prefix-Tree Decoding for Predicting Mass Spectra from Molecules},
  booktitle    = {Advances in Neural Information Processing Systems 36: Annual Conference
                  on Neural Information Processing Systems 2023, NeurIPS 2023, New Orleans,
                  LA, USA, December 10 - 16, 2023},
  year         = {2023},
  url          = {http://papers.nips.cc/paper\_files/paper/2023/hash/97d596ca21d0751ba2c633bad696cf7f-Abstract-Conference.html},
  bibsource    = {dblp computer science bibliography, https://dblp.org}
}

@Article{wang2016sharing,
author={Wang, Mingxun
and Carver, Jeremy J.
and Phelan, Vanessa V.
and Sanchez, Laura M.
and Garg, Neha
and Peng, Yao
and Nguyen, Don Duy
and Watrous, Jeramie
and Kapono, Clifford A.
and Luzzatto-Knaan, Tal
and Porto, Carla
and Bouslimani, Amina
and Melnik, Alexey V.
and Meehan, Michael J.
and Liu, Wei-Ting
and Cr{\"u}semann, Max
and Boudreau, Paul D.
and Esquenazi, Eduardo
and Sandoval-Calder{\'o}n, Mario
and Kersten, Roland D.
and Pace, Laura A.
and Quinn, Robert A.
and Duncan, Katherine R.
and Hsu, Cheng-Chih
and Floros, Dimitrios J.
and Gavilan, Ronnie G.
and Kleigrewe, Karin
and Northen, Trent
and Dutton, Rachel J.
and Parrot, Delphine
and Carlson, Erin E.
and Aigle, Bertrand
and Michelsen, Charlotte F.
and Jelsbak, Lars
and Sohlenkamp, Christian
and Pevzner, Pavel
and Edlund, Anna
and McLean, Jeffrey
and Piel, J{\"o}rn
and Murphy, Brian T.
and Gerwick, Lena
and Liaw, Chih-Chuang
and Yang, Yu-Liang
and Humpf, Hans-Ulrich
and Maansson, Maria
and Keyzers, Robert A.
and Sims, Amy C.
and Johnson, Andrew R.
and Sidebottom, Ashley M.
and Sedio, Brian E.
and Klitgaard, Andreas
and Larson, Charles B.
and Boya P, Cristopher A.
and Torres-Mendoza, Daniel
and Gonzalez, David J.
and Silva, Denise B.
and Marques, Lucas M.
and Demarque, Daniel P.
and Pociute, Egle
and O'Neill, Ellis C.
and Briand, Enora
and Helfrich, Eric J. N.
and Granatosky, Eve A.
and Glukhov, Evgenia
and Ryffel, Florian
and Houson, Hailey
and Mohimani, Hosein
and Kharbush, Jenan J.
and Zeng, Yi
and Vorholt, Julia A.
and Kurita, Kenji L.
and Charusanti, Pep
and McPhail, Kerry L.
and Nielsen, Kristian Fog
and Vuong, Lisa
and Elfeki, Maryam
and Traxler, Matthew F.
and Engene, Niclas
and Koyama, Nobuhiro
and Vining, Oliver B.
and Baric, Ralph
and Silva, Ricardo R.
and Mascuch, Samantha J.
and Tomasi, Sophie
and Jenkins, Stefan
and Macherla, Venkat
and Hoffman, Thomas
and Agarwal, Vinayak
and Williams, Philip G.
and Dai, Jingqui
and Neupane, Ram
and Gurr, Joshua
and Rodr{\'i}guez, Andr{\'e}s M. C.
and Lamsa, Anne
and Zhang, Chen
and Dorrestein, Kathleen
and Duggan, Brendan M.
and Almaliti, Jehad
and Allard, Pierre-Marie
and Phapale, Prasad
and Nothias, Louis-Felix
and Alexandrov, Theodore
and Litaudon, Marc
and Wolfender, Jean-Luc
and Kyle, Jennifer E.
and Metz, Thomas O.
and Peryea, Tyler
and Nguyen, Dac-Trung
and VanLeer, Danielle
and Shinn, Paul
and Jadhav, Ajit
and M{\"u}ller, Rolf
and Waters, Katrina M.
and Shi, Wenyuan
and Liu, Xueting
and Zhang, Lixin
and Knight, Rob
and Jensen, Paul R.
and Palsson, Bernhard {\O}
and Pogliano, Kit
and Linington, Roger G.
and Guti{\'e}rrez, Marcelino
and Lopes, Norberto P.
and Gerwick, William H.
and Moore, Bradley S.
and Dorrestein, Pieter C.
and Bandeira, Nuno},
title={Sharing and community curation of mass spectrometry data with Global Natural Products Social Molecular Networking},
journal={Nature Biotechnology},
year={2016},
month=aug,
day={01},
volume={34},
number={8},
pages={828-837},
issn={1546-1696},
doi={10.1038/nbt.3597},
url={https://doi.org/10.1038/nbt.3597}
}

@misc{che2026comparative,
      title={Comparative Analysis of Formula and Structure Prediction from Tandem Mass Spectra}, 
      author={Xujun Che and Xiuxia Du and Depeng Xu},
      year={2026},
      eprint={2601.00941},
      archivePrefix={arXiv},
      primaryClass={q-bio.QM},
      url={https://arxiv.org/abs/2601.00941}, 
}

@misc{zhong2026flexms,
      title={{FlexMS} is a flexible framework for benchmarking deep learning-based mass spectrum prediction tools in metabolomics}, 
      author={Yunhua Zhong and Yixuan Tang and Yifan Li and Jie Yang and Pan Liu and Jun Xia},
      year={2026},
      eprint={2602.22822},
      archivePrefix={arXiv},
      primaryClass={cs.AI},
      url={https://arxiv.org/abs/2602.22822}, 
}

@article{pollmann2026bridging,
author = {Julian Pollmann  and Roman Bushuiev  and Anton Bushuiev  and Tomáš Pluskal  and Florian Huber },
title = {Bridging MS2 Spectra and Chemical Space: Advances in Spectral Similarity, Molecular Retrieval, and De Novo Structure Discovery},
journal = {ChemRxiv},
volume = {2026},
number = {0309},
pages = {},
year = {2026},
doi = {10.26434/chemrxiv.15000536/v2},
URL = {https://chemrxiv.org/doi/abs/10.26434/chemrxiv.15000536/v2},
eprint = {https://chemrxiv.org/doi/pdf/10.26434/chemrxiv.15000536/v2}}

@article {skrinjar2025have,
	author = {{\v S}krinjar, Peter and Eberhardt, J{\'e}r{\^o}me and Tauriello, Gerardo and Schwede, Torsten and Durairaj, Janani},
	title = {Have protein-ligand cofolding methods moved beyond memorisation?},
	elocation-id = {2025.02.03.636309},
	year = {2025},
	doi = {10.1101/2025.02.03.636309},
	publisher = {Cold Spring Harbor Laboratory},
	URL = {https://www.biorxiv.org/content/early/2025/08/04/2025.02.03.636309},
	eprint = {https://www.biorxiv.org/content/early/2025/08/04/2025.02.03.636309.full.pdf},
	journal = {bioRxiv}
}

@article{ahmad2022chemberta2,
  author       = {Walid Ahmad and
                  Elana Simon and
                  Seyone Chithrananda and
                  Gabriel Grand and
                  Bharath Ramsundar},
  title        = {{ChemBERTa-2}: Towards Chemical Foundation Models},
  journal      = {CoRR},
  volume       = {abs/2209.01712},
  year         = {2022},
  url          = {https://doi.org/10.48550/arXiv.2209.01712},
  doi          = {10.48550/ARXIV.2209.01712},
  eprinttype   = {arXiv},
  eprint       = {2209.01712},
  bibsource    = {dblp computer science bibliography, https://dblp.org}
}

@article{bushuiev2024revealing,
  author       = {Anton Bushuiev and
                  Roman Bushuiev and
                  Jir{\'{\i}} Sedl{\'{a}}r and
                  Tom{\'{a}}s Pluskal and
                  Jir{\'{\i}} Damborsk{\'{y}} and
                  Stanislav Mazurenko and
                  Josef Sivic},
  title        = {Revealing data leakage in protein interaction benchmarks},
  journal      = {CoRR},
  volume       = {abs/2404.10457},
  year         = {2024},
  url          = {https://doi.org/10.48550/arXiv.2404.10457},
  doi          = {10.48550/ARXIV.2404.10457},
  eprinttype    = {arXiv},
  eprint       = {2404.10457},
  bibsource    = {dblp computer science bibliography, https://dblp.org}
}

@Article{kretschmer2025coverage,
author={Kretschmer, Fleming
and Seipp, Jan
and Ludwig, Marcus
and Klau, Gunnar W.
and B{\"o}cker, Sebastian},
title={Coverage bias in small molecule machine learning},
journal={Nature Communications},
year={2025},
month=jan,
day={09},
volume={16},
number={1},
pages={554},
issn={2041-1723},
doi={10.1038/s41467-024-55462-w},
url={https://doi.org/10.1038/s41467-024-55462-w}
}

@Article{mysinger2012directory,
author={Mysinger, Michael M.
and Carchia, Michael
and Irwin, John J.
and Shoichet, Brian K.},
title={Directory of Useful Decoys, Enhanced ({DUD-E}): Better Ligands and Decoys for Better Benchmarking},
journal={Journal of Medicinal Chemistry},
year={2012},
month=jul,
day={26},
publisher={American Chemical Society},
volume={55},
number={14},
pages={6582-6594},
issn={0022-2623},
doi={10.1021/jm300687e},
url={https://doi.org/10.1021/jm300687e}
}

@misc{macdiarmid2025natural,
      title={Natural Emergent Misalignment from Reward Hacking in Production {RL}}, 
      author={Monte MacDiarmid and Benjamin Wright and Jonathan Uesato and Joe Benton and Jon Kutasov and Sara Price and Naia Bouscal and Sam Bowman and Trenton Bricken and Alex Cloud and Carson Denison and Johannes Gasteiger and Ryan Greenblatt and Jan Leike and Jack Lindsey and Vlad Mikulik and Ethan Perez and Alex Rodrigues and Drake Thomas and Albert Webson and Daniel Ziegler and Evan Hubinger},
      year={2025},
      eprint={2511.18397},
      archivePrefix={arXiv},
      primaryClass={cs.AI},
      url={https://arxiv.org/abs/2511.18397}, 
}

@Inbook{goodhart1984problems,
author="Goodhart, C. A. E.",
title="Problems of Monetary Management: The UK Experience",
bookTitle="Monetary Theory and Practice: The UK Experience",
year="1984",
publisher="Macmillan Education UK",
address="London",
pages="91--121",
isbn="978-1-349-17295-5",
doi="10.1007/978-1-349-17295-5_4",
url="https://doi.org/10.1007/978-1-349-17295-5_4"
}

@Article{duhrkop2021systematic,
author={D{\"u}hrkop, Kai
and Nothias, Louis-F{\'e}lix
and Fleischauer, Markus
and Reher, Raphael
and Ludwig, Marcus
and Hoffmann, Martin A.
and Petras, Daniel
and Gerwick, William H.
and Rousu, Juho
and Dorrestein, Pieter C.
and B{\"o}cker, Sebastian},
title={Systematic classification of unknown metabolites using high-resolution fragmentation mass spectra},
journal={Nature Biotechnology},
year={2021},
month=apr,
day={01},
volume={39},
number={4},
pages={462-471},
issn={1546-1696},
doi={10.1038/s41587-020-0740-8},
url={https://doi.org/10.1038/s41587-020-0740-8}
}

@article{thomas2022reliance,
title = {Reliance on metrics is a fundamental challenge for {AI}},
journal = {Patterns},
volume = {3},
number = {5},
pages = {100476},
year = {2022},
issn = {2666-3899},
doi = {10.1016/j.patter.2022.100476},
url = {https://www.sciencedirect.com/science/article/pii/S2666389922000563},
author = {Rachel L. Thomas and David Uminsky}
}

@inproceedings{liao2021are,
  author       = {Thomas Liao and
                  Rohan Taori and
                  Inioluwa Deborah Raji and
                  Ludwig Schmidt},
  editor       = {Joaquin Vanschoren and
                  Sai{-}Kit Yeung},
  title        = {Are We Learning Yet? {A} Meta Review of Evaluation Failures Across
                  Machine Learning},
  booktitle    = {Proceedings of the Neural Information Processing Systems Track on
                  Datasets and Benchmarks 1, NeurIPS Datasets and Benchmarks 2021, December
                  2021, virtual},
  year         = {2021},
  url          = {https://datasets-benchmarks-proceedings.neurips.cc/paper/2021/hash/757b505cfd34c64c85ca5b5690ee5293-Abstract-round2.html},
  bibsource    = {dblp computer science bibliography, https://dblp.org}
}

@inproceedings{raji2021ai,
  author       = {Inioluwa Deborah Raji and
                  Emily Denton and
                  Emily M. Bender and
                  Alex Hanna and
                  Amandalynne Paullada},
  editor       = {Joaquin Vanschoren and
                  Sai{-}Kit Yeung},
  title        = {{AI} and the Everything in the Whole Wide World Benchmark},
  booktitle    = {Proceedings of the Neural Information Processing Systems Track on
                  Datasets and Benchmarks 1, NeurIPS Datasets and Benchmarks 2021, December
                  2021, virtual},
  year         = {2021},
  url          = {https://datasets-benchmarks-proceedings.neurips.cc/paper/2021/hash/084b6fbb10729ed4da8c3d3f5a3ae7c9-Abstract-round2.html},
  bibsource    = {dblp computer science bibliography, https://dblp.org}
}

@inproceedings{recht2019do,
  author       = {Benjamin Recht and
                  Rebecca Roelofs and
                  Ludwig Schmidt and
                  Vaishaal Shankar},
  editor       = {Kamalika Chaudhuri and
                  Ruslan Salakhutdinov},
  title        = {Do {ImageNet} Classifiers Generalize to {ImageNet}?},
  booktitle    = {Proceedings of the 36th International Conference on Machine Learning,
                  {ICML} 2019, 9-15 June 2019, Long Beach, California, {USA}},
  series       = {Proceedings of Machine Learning Research},
  volume       = {97},
  pages        = {5389--5400},
  publisher    = {{PMLR}},
  year         = {2019},
  url          = {http://proceedings.mlr.press/v97/recht19a.html},
  bibsource    = {dblp computer science bibliography, https://dblp.org}
}

@ARTICLE{Duhrkop2019sirius4,
  title     = "{SIRIUS} 4: a rapid tool for turning tandem mass spectra into
               metabolite structure information",
  author    = "Dührkop, Kai and Fleischauer, Markus and Ludwig, Marcus and
               Aksenov, Alexander A and Melnik, Alexey V and Meusel, Marvin and
               Dorrestein, Pieter C and Rousu, Juho and Böcker, Sebastian",
  journal   = "Nature Methods",
  publisher = "Springer Science and Business Media LLC",
  volume    =  16,
  number    =  4,
  pages     = "299--302",
  month     =  apr,
  year      =  2019,
  language  = "en"
}

@ARTICLE{Butler2023ms2mol,
  title    = "{MS2Mol}: A transformer model for illuminating dark chemical space
              from mass spectra",
  author   = "Butler, Thomas and Frandsen, Abraham and Lightheart, Rose and
              Bargh, Brian and Kerby, Thomas and West, Kiana and Davison, Joseph
              and Taylor, James and Krettler, Christoph and Bollerman, T J and
              Voronov, Gennady and Moon, Kevin and Kind, Tobias and Dorrestein,
              Pieter and Allen, August and Colluru, Viswa and Healey, David",
  journal  = "ChemRxiv",
  month    =  sep,
  year     =  2023,
  language = "en"
}

@ARTICLE{Litsa2023deep,
  title     = "An end-to-end deep learning framework for translating mass
               spectra to de-novo molecules",
  author    = "Litsa, Eleni E and Chenthamarakshan, Vijil and Das, Payel and
               Kavraki, Lydia E",
  journal   = "Communications Chemistry",
  publisher = "Springer Science and Business Media LLC",
  volume    =  6,
  number    =  1,
  pages     = "1--12",
  month     =  jun,
  year      =  2023,
  language  = "en"
}

@ARTICLE{Le2020neuraldecipher,
  title     = "Neuraldecipher - reverse-engineering extended-connectivity
               fingerprints ({ECFPs}) to their molecular structures",
  author    = "Le, Tuan and Winter, Robin and Noé, Frank and Clevert, Djork-Arné",
  journal   = "Chemical Science",
  publisher = "The Royal Society of Chemistry",
  volume    =  11,
  number    =  38,
  pages     = "10378--10389",
  month     =  sep,
  year      =  2020,
  language  = "en"
}

@Article{hoffmann2023mad,
AUTHOR = {Hoffmann, Martin A. and Kretschmer, Fleming and Ludwig, Marcus and Böcker, Sebastian},
TITLE = {{MAD HATTER} Correctly Annotates 98\% of Small Molecule Tandem Mass Spectra Searching in {PubChem}},
JOURNAL = {Metabolites},
VOLUME = {13},
YEAR = {2023},
NUMBER = {3},
ARTICLE-NUMBER = {314},
URL = {https://www.mdpi.com/2218-1989/13/3/314},
PubMedID = {36984753},
ISSN = {2218-1989},
DOI = {10.3390/metabo13030314}
}

@misc{dewaele2026small,
      title={Small molecule retrieval from tandem mass spectrometry: what are we optimizing for?}, 
      author={Gaetan De Waele and Marek Wydmuch and Krzysztof Dembczyński and Wojciech Kotłowski and Willem Waegeman},
      year={2026},
      eprint={2602.16507},
      archivePrefix={arXiv},
      primaryClass={cs.LG},
      url={https://arxiv.org/abs/2602.16507}, 
      doi={https://doi.org/10.48550/arXiv.2602.16507}
}

@inproceedings{bohde2025diffms,
  title={{DiffMS}: Diffusion Generation of Molecules Conditioned on Mass Spectra},
  author={Bohde, Montgomery and Manjrekar, Mrunali and Wang, Runzhong and Ji, Shuiwang and Coley, Connor W},
  booktitle={International Conference on Machine Learning},
  pages={4737--4756},
  year={2025},
  organization={PMLR}
}

@book{breitmaier2002structure,
  title={Structure elucidation by NMR in organic chemistry: a practical guide},
  author={Breitmaier, Eberhard},
  year={2002},
  publisher={John Wiley \& Sons},
  doi={10.1002/0470853069}
}

@article{iceberg2025,
  title={Neural spectral prediction for structure elucidation with tandem mass spectrometry},
  author={Wang, Runzhong and Manjrekar, Mrunali and Mahjour, Babak and Avila-Pacheco, Julian and Provenzano, Joules and Reynolds, Erin and Lederbauer, Magdalena and Mashin, Eivgeni and Goldman, Samuel and Wang, Mingxun and others},
  journal={bioRxiv},
  year={2025}
}

@article{mismetti2025ttt,
  title={Test-Time Tuned Language Models Enable End-to-end De Novo Molecular Structure Generation from {MS/MS} Spectra},
  author={Mismetti, Laura and Alberts, Marvin and Krause, Andreas and Graziani, Mara},
  journal={arXiv preprint arXiv:2510.23746},
  year={2025}
}

@Article{atanasov2021natural,
author={Atanasov, Atanas G.
and Zotchev, Sergey B.
and Dirsch, Verena M.
and Orhan, Ilkay Erdogan
and Banach, Maciej
and Rollinger, Judith M.
and Barreca, Davide
and Weckwerth, Wolfram
and Bauer, Rudolf
and Bayer, Edward A.
and Majeed, Muhammed
and Bishayee, Anupam
and Bochkov, Valery
and Bonn, G{\"u}nther K.
and Braidy, Nady
and Bucar, Franz
and Cifuentes, Alejandro
and D'Onofrio, Grazia
and Bodkin, Michael
and Diederich, Marc
and Dinkova-Kostova, Albena T.
and Efferth, Thomas
and El Bairi, Khalid
and Arkells, Nicolas
and Fan, Tai-Ping
and Fiebich, Bernd L.
and Freissmuth, Michael
and Georgiev, Milen I.
and Gibbons, Simon
and Godfrey, Keith M.
and Gruber, Christian W.
and Heer, Jag
and Huber, Lukas A.
and Ibanez, Elena
and Kijjoa, Anake
and Kiss, Anna K.
and Lu, Aiping
and Macias, Francisco A.
and Miller, Mark J. S.
and Mocan, Andrei
and M{\"u}ller, Rolf
and Nicoletti, Ferdinando
and Perry, George
and Pittal{\`a}, Valeria
and Rastrelli, Luca
and Ristow, Michael
and Russo, Gian Luigi
and Silva, Ana Sanches
and Schuster, Daniela
and Sheridan, Helen
and Skalicka-Wo{\'{z}}niak, Krystyna
and Skaltsounis, Leandros
and Sobarzo-S{\'a}nchez, Eduardo
and Bredt, David S.
and Stuppner, Hermann
and Sureda, Antoni
and Tzvetkov, Nikolay T.
and Vacca, Rosa Anna
and Aggarwal, Bharat B.
and Battino, Maurizio
and Giampieri, Francesca
and Wink, Michael
and Wolfender, Jean-Luc
and Xiao, Jianbo
and Yeung, Andy Wai Kan
and Lizard, G{\'e}rard
and Popp, Michael A.
and Heinrich, Michael
and Berindan-Neagoe, Ioana
and Stadler, Marc
and Daglia, Maria
and Verpoorte, Robert
and Supuran, Claudiu T.
and the International Natural Product Sciences Taskforce},
title={Natural products in drug discovery: advances and opportunities},
journal={Nature Reviews Drug Discovery},
year={2021},
month=mar,
day={01},
volume={20},
number={3},
pages={200-216},
issn={1474-1784},
doi={10.1038/s41573-020-00114-z},
url={https://doi.org/10.1038/s41573-020-00114-z}
}

@Article{stienstra2025structure,
author={Stienstra, Cailum M. K.
and Song, Jaegun
and Healey, David
and Voronov, Gennady
and Gardner, Eric
and Patel, Abhishek
and Macherla, Venkat
and Krettler, Christoph A.
and Kind, Tobias
and Dorrestein, Pieter C.
and Domingo-Fern{\'a}ndez, Daniel},
title={Structure characterization with {NMR} molecular networking},
journal={Communications Chemistry},
year={2025},
month=dec,
day={17},
volume={9},
number={1},
pages={28},
issn={2399-3669},
doi={10.1038/s42004-025-01839-x},
url={https://doi.org/10.1038/s42004-025-01839-x}
}

@inproceedings{alberts2024multimodal,
 author = {Alberts, Marvin and Schilter, Oliver and Zipoli, Federico and Hartrampf, Nina and Laino, Teodoro},
 booktitle = {Advances in Neural Information Processing Systems},
 doi = {10.52202/079017-3996},
 editor = {A. Globerson and L. Mackey and D. Belgrave and A. Fan and U. Paquet and J. Tomczak and C. Zhang},
 pages = {125780--125808},
 publisher = {Curran Associates, Inc.},
 title = {Unraveling Molecular Structure: A Multimodal Spectroscopic Dataset for Chemistry},
 url = {https://proceedings.neurips.cc/paper_files/paper/2024/file/e38e60b33bb2c6993e0865160cdb5cf1-Paper-Datasets_and_Benchmarks_Track.pdf},
 volume = {37},
 year = {2024}
}

@ARTICLE{Goldman2024mistcf,
  title     = "{MIST}-{CF}: Chemical Formula Inference from tandem mass spectra",
  author    = "Goldman, Samuel and Xin, Jiayi and Provenzano, Joules and Coley,
               Connor W",
  journal   = "Journal of Chemical Information and Modeling",
  publisher = "American Chemical Society",
  volume    =  64,
  number    =  7,
  pages     = "2421--2431",
  month     =  apr,
  year      =  2024,
  language  = "en"
}

@article{grulke2019dsstox,
  title={{EPA}'s {DSSTox} database: History of development of a curated chemistry resource supporting computational toxicology research},
  author={Grulke, Christopher M and Williams, Antony J and Thillanadarajah, Inthirany and Richard, Ann M},
  journal={Computational Toxicology},
  volume={12},
  pages={100096},
  year={2019},
  doi={10.1016/j.comtox.2019.100096}
}

@article{wishart2022hmdb,
  title={{HMDB} 5.0: the Human Metabolome Database for 2022},
  author={Wishart, David S and Guo, AnChi and Oler, Eponine and Wang, Fei and Anjum, Afia and Peters, Harrison and Diber, Raynard and Liang, Zinat and Gautam, Vasuk and Wishart, Dan and others},
  journal={Nucleic Acids Research},
  volume={50},
  number={D1},
  pages={D622--D631},
  year={2022},
  doi={10.1093/nar/gkab1062}
}

@article{sorokina2021coconut,
  title={{COCONUT} online: Collection of Open Natural Products database},
  author={Sorokina, Maria and Merseburger, Peter and Rajan, Kohulan and Yirik, Mehmet Aziz and Steinbeck, Christoph},
  journal={Journal of Cheminformatics},
  volume={13},
  number={1},
  pages={2},
  year={2021},
  doi={10.1186/s13321-020-00478-9}
}

@article{polykovskiy2020moses,
  title={Molecular Sets ({MOSES}): A Benchmarking Platform for Molecular Generation Models},
  author={Polykovskiy, Daniil and Zhebrak, Alexander and Sanchez-Lengeling, Benjamin and Golovanov, Sergey and Tatanov, Oktai and Belyaev, Stanislav and Kurbanov, Rauf and Artamonov, Aleksey and Aladinskiy, Vladimir and Veselov, Mark and Kadurin, Artur and Johansson, Simon and Chen, Hongming and Nikolenko, Sergey and Aspuru-Guzik, Al{\'a}n and Zhavoronkov, Alex},
  journal={Frontiers in Pharmacology},
  volume={11},
  pages={565644},
  year={2020},
  doi={10.3389/fphar.2020.565644}
}

@article{irwin2020zinc20,
title={{ZINC}20—A Free Ultralarge-Scale Chemical Database for Ligand Discovery},
author={Irwin, John J. and Tang, Khanh G. and Young, Jennifer and Dandarchuluun, Chinzorig and Wong, Benjamin R. and Khurelbaatar, Munkhzul and Moroz, Yurii S. and Mayfield, John and Sayle, Roger A.},
journal={Journal of Chemical Information and Modeling},
volume={60},
number={12},
pages={6065–6073},
year={2020},
doi={10.1021/acs.jcim.0c00675}
}

@article{chambers2013unichem,
  title={{UniChem}: a unified chemical structure cross-referencing and identifier tracking system},
  author={Chambers, Jon and Davies, Mark and Gaulton, Anna and Hersey, Anne and Velankar, Sameer and Petryszak, Robert and Hastings, Janna and Overington, John P and Steinbeck, Christoph and Leach, Andrew R},
  journal={Journal of Cheminformatics},
  volume={5},
  number={1},
  pages={3},
  year={2013},
  doi={10.1186/1758-2946-5-3}
}

@inproceedings{wang2025madgen,
  title={{MADGEN}: Mass-Spec attends to De Novo Molecular generation},
  author={Wang, Yinkai and Chen, Xiaohui and Liu, Liping and Hassoun, Soha},
  booktitle={The Thirteenth International Conference on Learning Representations},
  year={2025},
  url={https://openreview.net/forum?id=78tc3EiUrN}
}

@inproceedings{han2025msbart,
  title={{MS-BART}: Unified Modeling of Mass Spectra and Molecules for Structure Elucidation},
  author={Han, Yang and Wang, Pengyu and Yu, Kai and Chen, Xin and Chen, Lu},
  booktitle={Advances in Neural Information Processing Systems},
  year={2025}
}

@article{stravs2022msnovelist,
  title={{MSNovelist}: de novo structure generation from mass spectra},
  author={Stravs, Michael A and D{\"u}hrkop, Kai and B{\"o}cker, Sebastian and Zamboni, Nicola},
  journal={Nature Methods},
  volume={19},
  number={7},
  pages={865--870},
  year={2022},
  doi={10.1038/s41592-022-01486-3}
}

@article{beck2024recent,
author = {Beck, Armen G. and Muhoberac, Matthew and Randolph, Caitlin E. and Beveridge, Connor H. and Wijewardhane, Prageeth R. and Kenttämaa, Hilkka I. and Chopra, Gaurav},
title = {Recent Developments in Machine Learning for Mass Spectrometry},
journal = {ACS Measurement Science Au},
volume = {4},
number = {3},
pages = {233-246},
year = {2024},
doi = {10.1021/acsmeasuresciau.3c00060},
URL = { https://doi.org/10.1021/acsmeasuresciau.3c00060},
eprint = { https://doi.org/10.1021/acsmeasuresciau.3c00060}
}

@Article{schneider2026de,
AUTHOR = {Schneider, Mark Yu. and Kholmanskikh, Daniil D. and Romanov, Kirill Ya. and Perekina, Elena A. and Nikolenko, Sergei A. and Lukin, Ruslan Yu. and Golov, Ivan V.},
TITLE = {De Novo Structure Prediction from Tandem Mass Spectra: Algorithms, Benchmarks, and Limitations},
JOURNAL = {Molecules},
VOLUME = {31},
YEAR = {2026},
NUMBER = {5},
ARTICLE-NUMBER = {769},
URL = {https://www.mdpi.com/1420-3049/31/5/769},
PubMedID = {41828754},
ISSN = {1420-3049},
DOI = {10.3390/molecules31050769}
}

@article{whalen2021navigating,
  title = {Navigating the pitfalls of applying machine learning in genomics},
  volume = {23},
  ISSN = {1471-0064},
  url = {http://dx.doi.org/10.1038/s41576-021-00434-9},
  DOI = {10.1038/s41576-021-00434-9},
  number = {3},
  journal = {Nature Reviews Genetics},
  publisher = {Springer Science and Business Media LLC},
  author = {Whalen,  Sean and Schreiber,  Jacob and Noble,  William S. and Pollard,  Katherine S.},
  year = {2021},
  month = Nov,
  pages = {169–181}
}

@article{ozcelik2025how,
  title = {How evaluation choices distort the outcome of generative drug discovery},
  volume = {17},
  ISSN = {1758-2946},
  url = {http://dx.doi.org/10.1186/s13321-025-01108-y},
  DOI = {10.1186/s13321-025-01108-y},
  number = {1},
  journal = {Journal of Cheminformatics},
  publisher = {Springer Science and Business Media LLC},
  author = {\"{O}z\c{c}elik,  Rıza and Grisoni,  Francesca},
  year = {2025},
  month = Nov 
}

@software{greg_landrum_2026_19922430,
  author       = {Greg Landrum and
                  Paolo Tosco and
                  Brian Kelley and
                  Ricardo Rodriguez and
                  David Cosgrove and
                  Riccardo Vianello and
                  sriniker and
                  Peter Gedeck and
                  Gareth Jones and
                  Eisuke Kawashima and
                  NadineSchneider and
                  Dan Nealschneider and
                  tadhurst-cdd and
                  Andrew Dalke and
                  Matt Swain and
                  Brian Cole and
                  Samo Turk and
                  Aleksandr Savelev and
                  Niels Maeder and
                  Yakov Pechersky and
                  Alain Vaucher and
                  Maciej Wójcikowski and
                  Rachel Walker and
                  Hussein Faara and
                  Ichiru Take and
                  Vincent F. Scalfani and
                  Daniel Probst and
                  Kazuya Ujihara and
                  Jeremy Monat and
                  Juuso Lehtivarjo},
  title        = {rdkit/rdkit: 2026\_03\_2 ({Q1} 2026) Release},
  month        = apr,
  year         = 2026,
  publisher    = {Zenodo},
  version      = {Release\_2026\_03\_2},
  doi          = {10.5281/zenodo.19922430},
  url          = {https://doi.org/10.5281/zenodo.19922430},
}

@article{kalia2025jestr,
  title = {{JESTR}: Joint Embedding Space Technique for Ranking Candidate Molecules for the Annotation of Untargeted Metabolomics Data},
  author = {Kalia, Apurva and Chen, Yan Zhou and Krishnan, Dilip and Hassoun, Soha},
  journal = {Bioinformatics},
  volume = {41},
  number = {7},
  pages = {btaf354},
  year = {2025},
  doi = {10.1093/bioinformatics/btaf354}
}

@article{chen2026mvp,
  title = {Learning from All Views: A Multiview Contrastive Framework for Metabolite Annotation},
  author = {Chen, Yan Zhou and Hassoun, Soha},
  journal = {Analytical Chemistry},
  year = {2026},
  doi = {10.1021/acs.analchem.5c05675}
}

@article{neo2025molforge,
  title = {One Small Step with Fingerprints, One Giant Leap for De Novo Molecule Generation from Mass Spectra},
  author = {Neo, Neng Kai Nigel and Jing, Lim and Preston, Ngoui Yong Zhau and Serene, Koh Xue Ting and Shen, Bingquan},
  journal = {arXiv preprint arXiv:2508.04180},
  year = {2025},
  doi = {10.48550/arXiv.2508.04180},
  note = {Accepted at AI4Mat-NeurIPS-2025 Workshop}
}

@article{bohde2026frigid_v2,
  title={{FRIGID}: Scaling Diffusion-Based Molecular Generation from Mass Spectra at Training and Inference Time},
  author={Bohde, Montgomery and Liu, Hongxuan and Manjrekar, Mrunali and Lederbauer, Magdalena and Ji, Shuiwang and Wang, Runzhong and Coley, Connor W},
  journal={arXiv preprint arXiv:2604.16648v2},
  year={2026},
  url={https://arxiv.org/abs/2604.16648v2}
}

@article{bohde2026frigid_v3,
  title={{FRIGID}: Scaling Diffusion-Based Molecular Generation from Mass Spectra at Training and Inference Time},
  author={Bohde, Montgomery and Liu, Hongxuan and Manjrekar, Mrunali and Lederbauer, Magdalena and Ji, Shuiwang and Wang, Runzhong and Coley, Connor W},
  journal={arXiv preprint arXiv:2604.16648},
  year={2026},
  url={https://arxiv.org/abs/2604.16648}
}

@mastersthesis{khoo2025data,
  author = {Khoo, Ling Min Serena},
  title = {A Data Attribution-Based Approach to Model Diagnosis in {LC-MS/MS} Structure Prediction},
  school = {Massachusetts Institute of Technology},
  year = {2025},
  type = {Master's thesis},
  url = {https://dspace.mit.edu/handle/1721.1/164644}
}

@article{khoo2026why,
  author = {Khoo, Ling Min Serena and Barzilay, Regina},
  title = {Why machine learning fails at mass spectrometry for small molecules},
  journal = {Nature Metabolism},
  volume = {8},
  number = {6},
  pages = {1247--1249},
  year = {2026},
  doi = {10.1038/s42255-026-01544-6},
  url = {https://www.nature.com/articles/s42255-026-01544-6}
}

@article{khoo2026nearest,
  author = {Khoo, Ling Min Serena},
  title = {Nearest-neighbour baselines for fingerprint prediction from {MS/MS} spectra under different assumptions},
  journal = {arXiv preprint arXiv:2610.02249},
  year = {2026},
  url = {https://arxiv.org/abs/2610.02249}
}

@misc{sdbs,
  author       = {{National Institute of Advanced Industrial Science and Technology (AIST)}},
  title        = {{Spectral Database for Organic Compounds, SDBS}},
  howpublished = {\url{https://sdbs.db.aist.go.jp/}},

  note         = {Accessed: 2026-10-11},
  urldate      = {2026-10-11}
}

@article{huang2006benchmarking,
  title={Benchmarking sets for molecular docking},
  author={Huang, Niu and Shoichet, Brian K and Irwin, John J},
  journal={Journal of Medicinal Chemistry},
  volume={49},
  number={23},
  pages={6789--6801},
  year={2006},
  publisher={ACS Publications},
  doi={10.1021/jm0608356}
}

@article{bushuiev2025dreams,
  title     = {Self-supervised learning of molecular representations from millions of tandem mass spectra using {DreaMS}},
  author    = {Bushuiev, Roman and Bushuiev, Anton and Samusevich, Raman and Brungs, Corinna and Sivic, Josef and Pluskal, Tom{\'a}{\v{s}}},
  journal   = {Nature Biotechnology},
  volume    = {44},
  pages     = {630--640},
  year      = {2026},
  month     = {April},
  doi       = {10.1038/s41587-025-02663-3},
  note      = {Published online May 23, 2025}
}

@article{irwin2022chemformer,
  title     = {{Chemformer}: a pre-trained transformer for computational chemistry},
  author    = {Irwin, Ross and Dimitriadis, Spyridon and He, Jiazhen and Bjerrum, Esben Jannik},
  journal   = {Machine Learning: Science and Technology},
  volume    = {3},
  number    = {1},
  pages     = {015022},
  year      = {2022},
  publisher = {IOP Publishing},
  doi       = {10.1088/2632-2153/ac3ffb}
}

@article {gupta2026confronting,
	author = {Gupta, Vishu and Skinnider, Michael A.},
	title = {Confronting spurious evaluations of computational methods in small molecule mass spectrometry},
	elocation-id = {2026.05.03.722532},
	year = {2026},
	doi = {10.64898/2026.05.03.722532},
	publisher = {Cold Spring Harbor Laboratory},
	URL = {https://www.biorxiv.org/content/early/2026/05/06/2026.05.03.722532},
	eprint = {https://www.biorxiv.org/content/early/2026/05/06/2026.05.03.722532.full.pdf},
	journal = {bioRxiv}
}

@article{noutahi2023gotta,
  author       = {Emmanuel Noutahi and
                  Cristian Gabellini and
                  Michael Craig and
                  Jonathan S. C. Lim and
                  Prudencio Tossou},
  title        = {Gotta be {SAFE:} {A} New Framework for Molecular Design},
  journal      = {CoRR},
  volume       = {abs/2310.10773},
  year         = {2023},
  url          = {https://doi.org/10.48550/arXiv.2310.10773},
  doi          = {10.48550/ARXIV.2310.10773},
  eprinttype   = {arXiv},
  eprint       = {2310.10773},
  bibsource    = {dblp computer science bibliography, https://dblp.org}
}

@article{manjrekar2026generative,
  author       = {Mrunali Manjrekar and
                  Runzhong Wang and
                  Samuel Goldman and
                  Jenna C. Fromer and
                  Connor W. Coley},
  title        = {Generative structural elucidation from mass spectra as an iterative
                  optimization problem},
  journal      = {CoRR},
  volume       = {abs/2602.07709},
  year         = {2026},
  url          = {https://doi.org/10.48550/arXiv.2602.07709},
  doi          = {10.48550/ARXIV.2602.07709},
  eprinttype   = {arXiv},
  eprint       = {2602.07709},
  bibsource    = {dblp computer science bibliography, https://dblp.org}
}

@article{zhao2026ms,
  title={{MS-GPT}: Rethinking {MS/MS} De Novo Structure Elucidation as Spectrum-Induced Posterior Querying of a Molecule-Language Model},
  author={Zhao, Xin and Liu, Yumin and Li, Zhuo and Zheng, Weichu and Zhu, Feng and Yang, Xiaokang and Jin, Yaohui and Xu, Yanyan},
  journal={arXiv preprint arXiv:2607.23607},
  year={2026}
}

@article{wang2026glacier,
  title={{GLACIER}: Rethinking Mass Spectrum Prediction as an Object Detection Problem},
  author={Wang, Rui-Xi and Wang, Runzhong and Coley, Connor W},
  journal={arXiv preprint arXiv:2606.29161},
  year={2026}
}

@inproceedings{Qin2026MSAnchorDN,
  title={{MSAnchor}: De Novo Molecular Generation from Mass Spectrometry Data with Anchor-Extended Molecular Scaffolds},
  author={Xiaohan Qin and Chao Wang and Zhen-Qiang Zhou and Linjiang Chen and Wenjie Du and Yang Wang},
  booktitle={AAAI Conference on Artificial Intelligence},
  year={2026},
  url={https://api.semanticscholar.org/CorpusID:286668073}
}







\clearpage
\appendix

\section{Technical Appendices and Supplementary Material}

\subsection{Experimental Details and Full Results of MolForge Decoder Pretraining}
\label{sec:appendix_contamination_details}

\subsubsection{Data Contamination Condition Definitions}

All structural comparisons in this section use ECFP4 Morgan fingerprints (radius\,2, 2048\,bits) and the MassSpecGym test set as the reference. We curate three decoder pretraining corpora at different scales---2.5M, 10M, and 50M molecules---with six variations of similarity-based exclusion criteria each\footnote{\url{https://huggingface.co/datasets/harrylau/ms-gpt-18}}, discussed in detail below. The 2.5M splits are drawn from the composite dataset of \textbf{DSSTox}~\citep{grulke2019dsstox}, \textbf{HMDB}~\citep{wishart2022hmdb}, \textbf{COCONUT}~\citep{sorokina2021coconut}, and \textbf{MOSES}~\citep{polykovskiy2020moses} commonly used in prior work~\citep{neo2025molforge, bohde2025diffms, wang2025madgen, han2025msbart}. The 10M and 50M corpora are randomly sampled from a combined pool of approximately 1B molecules from \textbf{ZINC20}~\citep{irwin2020zinc20} and \textbf{UniChem}~\citep{chambers2013unichem}. For each size, we curate six dataset variants according to the following criteria:

\begin{itemize}[leftmargin=*]
    \item \textbf{\textsc{exclude\_exact}:} For each target size, we randomly draw molecules from the corresponding source corpus until obtaining 2.5M, 10M, or 50M molecules whose connectivity-layer InChIKey (first 14 characters, i.e., the 2D structure hash) does not match any MassSpecGym test-set molecule. Because the first InChIKey block ignores stereochemical layers, this criterion also excludes all stereoisomers of test molecules.
    \item \textbf{\textsc{exclude\_tani080}, \textsc{exclude\_tani070}, and \textsc{exclude\_tani060}:} Starting from the same source order, we replace all molecules whose maximum fingerprint Tanimoto similarity to the test set is greater than or equal to the specified threshold with additional molecules from the same source corpus that satisfy the threshold. We choose Tanimoto similarity for this experiment because MCES distance is expensive and slow to compute at scale. We expect that either metric should be sufficient for this experiment. For all pretraining molecules $x$ and test molecules $\mathcal{T}_{\mathrm{test}}$, the retained set satisfies $\max_{y \in \mathcal{T}_{\mathrm{test}}} \mathrm{Tan}(\mathrm{ECFP4}(x), \mathrm{ECFP4}(y)) < \theta$, with $\theta \in \{0.80,0.70,0.60\}$. 
    \item \textbf{\textsc{include\_half}:} We construct a partially contaminated condition by starting from \textsc{exclude\_exact} and replacing an equal number of randomly selected non-test molecules with 50\% of the unique MassSpecGym test-set molecules. This isolates the effect of direct test-structure exposure while preserving the total number of decoder pretraining molecules.
    \item \textbf{\textsc{include\_all}:} We construct a fully contaminated condition analogously, replacing non-test molecules in \textsc{exclude\_exact} with all unique MassSpecGym test-set molecules. This provides an upper-bound direct-leakage condition under otherwise identical corpus size and training settings.
\end{itemize}

\begin{table}[htbp!]
\centering
\caption{\textbf{Dataset Construction Audit for the Six Leakage Control Variants at Each Decoder Pretraining Scale.} Entries are the number of molecules removed relative to the size-matched \textsc{exclude\_exact} construction for Tanimoto-filtered variants, or the number of MassSpecGym test-set molecules inserted into the \textsc{include} variants. Note that the 2.5M dataset is drawn from a different source corpus from the 10M and 50M datasets, and has a much higher concentration of close structure analogs to MassSpecGym test structures. This may affect the interpretation of Figure~\ref{fig:decoder_memorization}c and Table~\ref{tab:contamination_full}, giving more weight to the 2.5M model.}
\label{tab:contamination_dataset_audit}
\vspace{5pt}
\small
\resizebox{\linewidth}{!}{%
\begin{tabular}{lcccccc}
\toprule
Target size & \textsc{exclude\_exact} & \textsc{exclude\_tani080} & \textsc{exclude\_tani070} & \textsc{exclude\_tani060} & \textsc{include\_half} & \textsc{include\_all} \\
\midrule
2.5M & 0 removed & 2,174 additionally removed & 5,939 additionally removed & 22,504 additionally removed & 1,585 inserted & 3,170 inserted \\
10M  & 0 removed & 559 additionally removed & 1,633 additionally removed & 6,495 additionally removed & 1,585 inserted & 3,170 inserted \\
50M  & 0 removed & 2,666 additionally removed & 7,946 additionally removed & 32,070 additionally removed & 1,585 inserted & 3,170 inserted \\
\bottomrule
\end{tabular}
}
\end{table}

\subsubsection{Molecular Decoder Training and Inference Configuration}

We train a separate SMILES decoder~\citep{zhao2026ms} for each dataset variant and evaluate its checkpoints at the three corpus sizes. Decoder training is varied while the spectrum encoder and evaluation protocol are held fixed. The decoder uses 12 Transformer layers with hidden size 768, 12 attention heads, dropout 0.1, and a maximum sequence length of 256, with character-level SMILES tokenization. It is conditioned on 4096-bit Morgan fingerprints and molecular formulas through a two-layer fingerprint encoder and cross-attention inserted after every two decoder layers. Models are trained with AdamW ($\beta_1=0.9$, $\beta_2=0.95$), token-level negative log-likelihood loss, an effective batch size of 1{,}024, weight decay 0.01, gradient clipping at norm 1.0, and bfloat16 automatic mixed precision. The learning rate warms up over 2{,}000 optimizer steps to $3\times10^{-4}$ and then follows a cosine decay schedule.

For the 2.5M, 10M, and 50M corpus sizes, respectively, we train all dataset variants for 15, 8, and 3 epochs. These schedules were chosen to ensure that models at each scale reach convergence under the fixed training budget. For each trained model, we select the checkpoint with the best validation performance and use that checkpoint for test evaluation. Training is run in distributed mode on a node with 172 CPU cores, 2048 GB of memory, and 8 NVIDIA RTX PRO 6000 Blackwell Server Edition GPUs; wall-clock training time ranges from approximately 1.5 to 6 hours per model, depending on the corpus size and filtering condition.

Inference is implemented in \url{https://github.com/harrylaucngd/msms-models}. Each trained decoder is paired with the same MIST \textit{4M-aug} checkpoint to predict molecular fingerprints from peak-formula--annotated spectra. No end-to-end joint fine-tuning is performed, as it was observed to degrade performance. For each spectrum, we construct 20 binary fingerprint queries by applying equally spaced thresholds within a density-calibrated interval to the MIST probabilities. The interval is calibrated on the MassSpecGym training set: its endpoints yield mean predicted on-bit counts equal to 1.5 and 0.95 times the mean reference on-bit count, respectively. We sample five SMILES sequences per query, giving 100 candidates per spectrum, with temperature 1.0, top-$k=50$, and top-$p=0.95$. Valid candidates are canonicalized, grouped by 2D InChIKey, filtered by the known molecular formula, and ranked by generation frequency; ties are broken by fingerprint Tanimoto similarity to the query fingerprints. We evaluate all 17{,}556 input spectra in the MassSpecGym \textit{de novo} test split and report spectrum-level averages. By contrast, earlier evaluations using MIST vDiffMS and the FRIGID evaluation in arXiv v2~\citep{bohde2026frigid_v2} excluded 474 spectra whose ground-truth molecules contain atom types outside the MIST-supported set $\{\mathrm{C}, \mathrm{O}, \mathrm{P}, \mathrm{N}, \mathrm{S}, \mathrm{Cl}, \mathrm{F}, \mathrm{H}\}$, reducing evaluation to 17{,}082 spectra. The corrected MIST + MolForge and FRIGID entries in Table~\ref{tab:massspecgym_leaderboard_full} are evaluated on all 17{,}556 test spectra. MCES is not computed in these experiments; reported metrics are top-1 accuracy (Acc@1) and top-10 accuracy (Acc@10), based on 2D InChIKey matching after tautomer canonicalization and removal of stereochemistry. Inference is also run in distributed mode on the same 172-core, 2048-GB, 8-GPU node; full 8-GPU inference over the benchmark takes approximately 1.5 hours per model.

For Figure~\ref{fig:decoder_memorization}a, we group the 17{,}556 test spectra into 2{,}997 compounds by InChIKey-14, average top-1 hits within each compound, and then average compounds within each similarity bin. Similarity is the maximum over the union of the encoder's MSG-4M augmentation molecules and MassSpecGym training molecules, and the decoder's 2.5M, 10M, and 50M training corpora. We use the molecular representations used during training and compute 95\% BCa intervals with 20{,}000 compound-level bootstrap resamples (seed 42); intervals are omitted for bins with zero empirical variance.

\subsubsection{Full Per-Condition Metric Table}

Table~\ref{tab:contamination_full} reports exact-match generation accuracy of the selected validation-best decoder checkpoint for every corpus size and leakage variant. Each cell gives Acc@1/Acc@10, rows are ordered by decoder pretraining scale, and columns are ordered from direct test-set inclusion to progressively stricter exclusion. The 2.5M decoder shows the worst-case performance gradient with respect to splitting criterion, motivating our recommendation that training datasets exclude, for example, structures with Tanimoto similarity $\geq 0.7$ to any test-set compound when reporting results about generalization.

\begin{table}[htbp!]
\vspace{-6pt}
\centering
\caption{\textbf{Exact-Match Accuracy on the MassSpecGym \textit{de novo} Challenge for Molecular Decoders Trained on All Corpus Sizes and Dataset Variants.} Each decoder is paired with MIST \textit{4M-aug}. Each entry reports Acc@1/Acc@10 for the checkpoint with the best validation performance.}
\label{tab:contamination_full}
\vspace{5pt}
\small
\resizebox{\linewidth}{!}{%
\begin{tabular}{lcccccc}
\toprule
Target size & \textsc{include\_all} & \textsc{include\_half} & \textsc{exclude\_exact} & \textsc{exclude\_tani080} &
\textsc{exclude\_tani070} & \textsc{exclude\_tani060} \\
\midrule
2.5M & 8.26/9.76\% & 7.16/8.51\% & 7.95/9.20\% & 7.60/9.02\% & 6.41/7.49\% & 6.04/7.89\% \\
10M  & 8.93/10.30\% & 9.06/10.51\% & 9.15/10.78\% & 8.75/10.01\% & 8.60/10.05\% & 8.19/9.68\% \\
50M  & 9.56/11.24\% & 9.57/11.33\% & 9.50/11.39\% & 9.28/10.75\% & 8.99/10.84\% & 8.87/10.38\% \\
\bottomrule
\end{tabular}
}
\vspace{-6pt}
\end{table}




\subsection{Experimental Details for Shared Components and Synthetic Data Leakage Study}
\label{sec:appendix_shared_components}

\subsubsection{GLACIER Training and Data Augmentation Configuration}
\label{sec:iceberg-augmentation}

GLACIER~\citep{wang2026glacier} is a spectral simulator that generates synthetic MS/MS spectra from molecular structures. A spectral simulator is also representative of a broader class of reused shared components in recent MS/MS work, including recent test-time tuning pipelines pretrained on simulated spectra that include ICEBERG generations~\citep{mismetti2025ttt}. Our implementation is based on the \url{https://github.com/coleygroup/ms-pred} codebase. We train two GLACIER variants under controlled conditions:

\begin{itemize}[leftmargin=*]
    \item \textbf{Leakage-free GLACIER:} Trained only on MassSpecGym training-set spectrum--structure pairs, with all test and validation molecules excluded. This model represents the data-safe simulator that respects the benchmark split.
    \item \textbf{Adversarial GLACIER:} Trained on all 231{,}104 MassSpecGym spectrum--structure pairs from the training, validation, and test sets combined, solely to construct an adversarial leakage scenario. This model is not intended as a deployable simulator; it represents a forward model that has directly learned spectrum--structure associations from the evaluation sets.
\end{itemize}

GLACIER jointly trains fragmentation and intensity prediction using a five-layer Graphormer with hidden size 256, a four-layer fragment decoder, and a three-layer intensity encoder and decoder. We use dropout 0.1, collision-energy, adduct, formula, element-group, and instrument embeddings, up to 200 fragmentation breakpoints and six broken bonds, and a 20\,ppm tolerance. Training uses AdamW with an effective batch size of 32, learning rate $4\times10^{-4}$, no weight decay, 2{,}000 warmup steps, and learning-rate decay by 0.9 every 5{,}000 steps thereafter. The objective combines cosine intensity loss with fragment supervision, using a base fragment-loss weight of 0.1; the MAGMa supervision weight is held fixed for 2{,}500 steps and then decayed by 0.9 every 2{,}500 steps. Models are trained for up to 200 epochs, with a minimum of 20 epochs and early-stopping patience of five epochs. For the adversarial model, the monitored validation fold overlaps with training and serves only as a stopping diagnostic. No contrastive fine-tuning is performed. Training and synthetic data generation are performed on one NVIDIA RTX PRO 6000 Blackwell Server Edition GPU. GLACIER is a relatively lightweight forward model in this setting; training each GLACIER model takes approximately 2 hours.

To examine leakage from simulator training and augmentation targets, we construct four synthetic augmentation corpora by crossing the two GLACIER variants with two molecule pools. The base pool is the official set of 4 million unlabeled molecules released with MassSpecGym~v1.0, which excludes molecules within MCES distance 2 of the test set. After excluding two structures whose formula counts exceed the MIST representation's supported range, this pool contains 3{,}922{,}525 molecules. For each molecule, five collision energies are sampled without replacement from 0 to 100\,eV in increments of 5\,eV and combined with two instrument types (Orbitrap and QTOF) and two adducts ($[\mathrm{M}+\mathrm{H}]^+$ and $[\mathrm{M}+\mathrm{Na}]^+$), yielding 20 spectra per molecule and 78{,}450{,}500 synthetic spectrum--molecule pairs. The test-spiked pool adds 2{,}998 MassSpecGym test structures under the same 20 conditions, adding 59{,}960 pairs for a total of 78{,}510{,}460.

The four corpora correspond to Table~\ref{tab:transitive_contamination} as follows: (1)~adversarial GLACIER on the test-spiked pool; (2)~adversarial GLACIER on the base pool; (3)~leakage-free GLACIER on the test-spiked pool; and (4)~leakage-free GLACIER on the base pool. Comparing Settings~3 and~4 measures the effect of adding test structures to the augmentation pool while holding the simulator fixed; comparing Settings~2 and~4 measures the effect of changing the simulator training corpus while holding the augmentation targets fixed. The adversarial training corpus also includes validation spectra and spectra with imputed collision energies, so this comparison includes these changes in simulator training coverage.

\subsubsection{MIST Training Configuration with GLACIER Augmentation Data}
\label{sec:mist}

We train four MIST encoders, each on the MassSpecGym training set combined with one of the four GLACIER augmentation corpora described above. Every model uses the same 194{,}119 experimental spectrum--molecule pairs, mixed with synthetic spectra throughout training. Each effective optimization batch contains 8{,}192 synthetic and 300 experimental pairs. We train for 478{,}850 optimizer steps, corresponding to 50 passes through the base synthetic corpus, and use the same training budget and mixing ratio for all four settings. The encoder in Setting~4 of Table~\ref{tab:transitive_contamination} is MIST \textit{4M-aug}.

In each case, the encoder predicts 4096-bit Morgan (ECFP4) fingerprints from peak-formula--annotated spectra, using positional--cosine formula embeddings, pairwise peak featurization, and up to 10 peaks. The model has four peak-attention layers with 16 heads, four fingerprint refinement layers, hidden size 1024, and spectra dropout 0.1. The loss combines cosine fingerprint loss, soft Tanimoto loss with weight 0.1, and iterative fingerprint supervision with weight 0.4; the MAGMa auxiliary loss of the original MIST is removed. Optimization uses Adam with peak learning rate $2\times10^{-4}$, 9{,}577 warmup steps followed by cosine decay, weight decay $10^{-6}$, gradient clipping at norm 5.0, bfloat16 mixed precision, and an exponential moving average of model weights with decay approximately 0.9975. On-the-fly augmentation is applied to half of the spectra, with peak-drop probability 0.1 and multiplicative intensity jitter with log-scale standard deviation 0.2. We evaluate every 9{,}577 steps and select the EMA checkpoint with the highest validation spectrum-to-fingerprint Tanimoto similarity. A scalar binarization threshold is calibrated on the validation set by matching the total predicted on-bit count to the total reference on-bit count, then fixed for test evaluation. After training, each model is benchmarked on the MassSpecGym test set using spectrum-to-fingerprint Tanimoto similarity and formula-constrained retrieval hit rate. All training and benchmarking are performed on a single NVIDIA RTX~PRO~6000 Blackwell Server Edition GPU, with training taking roughly 12 hours per model.

The complete training implementation is available in \url{https://github.com/coleygroup/MIST}. Building on the original \url{https://github.com/samgoldman97/mist} implementation, this updated repository provides a modern training framework with native distributed data parallel (DDP) support and integrated workflows for synthetic spectrum generation and augmented training. We recommend this repository for future MIST training.

\subsubsection{MIST-CF Formula Annotation Components}
\label{app:mist-cf}

MIST-CF~\citep{Goldman2024mistcf} is a machine learning model that predicts which molecular formula best explains an MS/MS spectrum from a set of candidate formulas. Because it can provide automated inference of formulas, it can serve as a formula annotator to pair with formula-aware architectures for evaluation in the mass-based MassSpecGym challenge tracks. As it may be reused as a fixed auxiliary model, its training provenance matters for the same reason as that of spectral simulators: a public checkpoint trained on overlapping data can transmit information into downstream evaluation even when the downstream split itself is clean.

\begin{table}[htbp!]
\vspace{-9pt}
\centering
\caption{\textbf{Performance of MIST-CF Trained and Evaluated on the MassSpecGym Benchmark.} Versions of MIST-CF with leakage of test spectra are colored in {\color{gray} gray}.}
\vspace{6pt}
\label{tab:mist-cf-accuracy}
{\small
\begin{tabular}{lcc}
\toprule
\multicolumn{1}{c}{MIST-CF training setup} & \multicolumn{2}{c}{Accuracy $\uparrow$} \\
\cmidrule(lr){2-3}
  & Top-1  & Top-5 \\
\midrule
Train samples & 50.5\% & 73.2\% \\
\color{gray} Both train and test samples & \color{gray} 59.0\% &  \color{gray} 80.5\% \\
\color{gray} Test samples & \color{gray} 66.5\% & \color{gray} 84\% \\
\bottomrule
\end{tabular}
}
\end{table}

We therefore retrain MIST-CF on the MassSpecGym dataset, as well as two contaminated splits, and report the top-1 and top-5 formula annotation accuracies in Table~\ref{tab:mist-cf-accuracy}.  MIST-CF is a formula Transformer that ranks candidate chemical formulas for an observed precursor mass using an energy-based approach: for each spectrum, a set of candidate formulas is enumerated via the SIRIUS decomposition algorithm, subformula assignments are precomputed for each candidate peak, and the model is trained to score the correct formula above a set of sampled decoys. The architecture uses 2 Transformer layers with hidden size 128, sinusoidal formula embeddings, dropout of 0.1, and embeds adduct type, instrument type, and neutral-loss fragment formulas as additional covariates; up to 20 subpeaks and 32 decoy formulas are used per training example. Models are trained with Adam (learning rate $4.5\times10^{-4}$, no weight decay) and a negative log-likelihood loss over the candidate set. The learning rate schedule applies a 100-step linear warmup followed by exponential decay at rate 0.88 every 5{,}000 steps. Training runs for up to 200 epochs with early stopping (patience 5) on validation loss, and the best-validation checkpoint is selected for inference. Training uses a single NVIDIA H100 GPU with 16 CPU workers and up to 200 GB of memory; all models (batch size 4) trained to convergence in approximately 7 hours over a maximum of 10 epochs. As earlier stated, we generate formula candidates from SIRIUS's mass decomposition algorithm, which takes the precursor $m/z$ and determines all plausible formulas. We use the SIRIUS default element tolerances, which include a maximum tolerance of 1 for each halogen element (F, Cl, Br, I). Since this protocol can result in potentially thousands of formula candidates for a single mass, we adopt the fast-filter model also developed by MIST-CF to filter candidates to a maximum of 256 candidates that MIST-CF itself needs to consider. This filter model is trained on all formulas present in the original MassSpecGym pretraining set, with all MassSpecGym test and validation formulas excluded. We release both the data-safe MIST-CF and the complementary fast-filter model in MassSpecGym~v1.5. The codebase also now supports evaluation of formula-aware models paired with formula annotators on the mass-based challenge. 

\begin{table*}[htbp!]
\caption{\textbf{Additional Results of MIST Retrieval with Top-5 MIST-CF Formula Predictions.} Retrieval performance of MIST on the MassSpecGym mass-based retrieval challenge, using top-5 formula predictions from three MIST-CF variants (standard, test-only, and train-and-test), aggregated via average or element-wise max across the five formula-conditioned fingerprint predictions. Inflated results are marked in \textcolor{gray}{gray}. The values in parentheses indicate 99.9\% BCa bootstrap confidence intervals (20,000 resamples).}
\vspace{-0.15in}
\label{tab:retrieval_leaderboard_mass_challenge_w_formula_addition}
\begin{center}
\renewcommand{\arraystretch}{1.15}
\small
\resizebox{\linewidth}{!}{%
\begin{tabular}{lllll}
\toprule
Method & Hit rate@1 $\uparrow$ & Hit rate@5 $\uparrow$ & Hit rate@20 $\uparrow$ & MCES@1 $\downarrow$ \\
\midrule
MIST (MIST-CF top-5 formula, element-wise max.) & 32.36 (31.23--33.48)\% & 47.73 (46.49--48.98)\% & 63.08 (61.93--64.28)\% &
15.24 (14.87--15.60) \\
\color{gray} MIST (MIST-CF test-only top-5 formula, average) & \color{gray} 34.44 (33.26--35.66)\% & \color{gray} 49.71
(48.48--50.96)\% & \color{gray} 65.24 (64.04--66.41)\% & \color{gray} 14.26 (13.91--14.61) \\
\color{gray} MIST (MIST-CF test-only top-5 formula, element-wise max.) & \color{gray} 35.42 (34.25--36.65)\% &
\color{gray} 49.85 (48.59--51.11)\% & \color{gray} 64.64 (63.47--65.84)\% & \color{gray} 14.02 (13.66--14.37) \\
\color{gray} MIST (MIST-CF train-and-test top-5 formula, average) & \color{gray} 34.61 (33.43--35.76)\% & \color{gray} 50.15
(48.92--51.39)\% & \color{gray} 65.16 (63.95--66.35)\% & \color{gray} 14.17 (13.82--14.51) \\
\color{gray} MIST (MIST-CF train-and-test top-5 formula, element-wise max.) & \color{gray} 34.55 (33.33--35.78)\% &
\color{gray} 49.67 (48.48--50.88)\% & \color{gray} 64.52 (63.39--65.67)\% & \color{gray} 14.28 (13.94--14.62) \\
\bottomrule
\end{tabular}
}
\end{center}
\end{table*}

We pair these models' predictions with MIST \textit{4M-aug}, a spectrum-to-fingerprint model, by either using the top-1 formula prediction or aggregating the fingerprints generated from the top-5 formula predictions by averaging or taking an element-wise maximum. We then perform retrieval (as measured by lowest Tanimoto distance to the fingerprint) on the mass-based candidates. Prefiltration of the mass-based candidates to only candidates that match the formula predictions is not applied here. The main results are reported in Table~\ref{tab:retrieval_leaderboard_mass_challenge_w_formula}; additional results based on MIST-CF's top-5 formula predictions are provided for comparison in Table~\ref{tab:retrieval_leaderboard_mass_challenge_w_formula_addition}.

\subsection{Nearest Neighbor Implementation and Reproduction}
\label{sec:appendix_nearest_neighbor}

\textbf{Evaluating nearest-neighbor baselines.}
\citet{khoo2026why} recently argued that nearest-neighbor search outperforms machine learning methods on mass spectrometry. However, their headline result rests on an evaluation where the nearest-neighbor baseline was scored on an easier subset of the test set than MIST,\footnote{A detailed analysis of the evaluation in \citet{khoo2026why}, along with code for reproduction, is available at \url{https://github.com/coleygroup/Audit_ML_MS_Analysis}. Note that on Sep 30, 2026, the commit history in the official repository was overwritten with updated methods and results that no longer match the published paper's table. For reproducibility, a snapshot of the original repository is preserved at \url{https://github.com/coleygroup/Audit_ML_MS_analysis/tree/original_repo_bkup}.} and where a method based on DreaMS~\citep{bushuiev2025dreams} outperforms vanilla nearest neighbor despite the titular claim. Nevertheless, we recognize nearest neighbor as a useful reference baseline and include it in the official MassSpecGym setting. Results in Table~\ref{tab:retrieval_leaderboard_full} show that while nearest neighbor is competitive with some earlier baselines, recent machine learning methods significantly outperform it. Notably, MIST \textit{4M-aug} outperforms nearest neighbor by a large margin, running counter to the claim in \citet{khoo2026why}: ``These findings contradict the hypothesis that low performance arises primarily from insufficient training coverage. Simply adding more data to existing models is unlikely to solve the problem.''

\textbf{Earlier baseline and split definitions.}
The baseline described by \citet{khoo2025data} returns the fingerprint of the training spectrum with the highest spectral similarity, without specifying a molecular-formula restriction. The supplementary methods of \citet{khoo2026why} explicitly restrict this search to training spectra with the query formula. However, in the accompanying implementation, queries with no same-formula training spectrum are skipped rather than counted as misses, restricting the evaluation to an easier subset of test queries.\footnote{See \href{https://github.com/coleygroup/Audit_ML_MS_analysis/blob/da67d9d0a2a01a11ea65b43bf12b2388f0b045f8/benchmarked_models/nearest_neighbour/02_compute_nn.py}{\texttt{02\_compute\_nn.py} at commit \texttt{da67d9d}}, where test entries without a matching training formula are omitted from evaluation.}

Separately, the custom MassSpecGym scaffold and random splits in \citet{khoo2026why} differ from the official benchmark splits and exhibit substantial data leakage. In the released dataset files, 95.3\% of random-split test spectra have the same molecule present in training, and the leakage rate is 14.3\% for the scaffold split, as confirmed by a follow-up study from the same author~\citep{khoo2026nearest}. Consequently, in addition to partial test-set coverage, \citet{khoo2026why} primarily evaluated training-set memorization rather than the out-of-distribution generalization targeted by most machine learning models (and targeted by the official MassSpecGym split). Moreover, the source of identical-molecule leakage in the scaffold split remains unclear: a standard scaffold split should assign identical compounds to the same partition, and no open implementation is available to understand the split creation.

\textbf{Test-set coverage under formula filtering.}
Table~\ref{tab:nn_formula_coverage} shows the resulting coverage. On the custom MassSpecGym scaffold split, the original code evaluates 6,571 of 16,042 test spectra (40.96\%) and omits 9,471. A score averaged over this subset cannot be compared directly with scores over the full test set. On the official MassSpecGym split, only 2,956 of 17,556 test spectra (16.84\%) have a same-formula training spectrum; applying the same restriction without a fallback would omit 83.16\% of the test set, making it inapplicable in a fair MassSpecGym setting.


\begin{table}[htbp!]
\centering
\caption{\textbf{Same-Formula Training Coverage of MassSpecGym Test Spectra.} Counts for the released scaffold and random splits are obtained from the released files; the final row uses the official benchmark split. Coverage is the fraction of test spectra with at least one same-formula training spectrum.}
\label{tab:nn_formula_coverage}
{\small
\begin{tabular}{lrrr}
\toprule
Split & Test spectra & With a formula match & Coverage \\
\midrule
Released scaffold & 16,042 & 6,571 & 40.96\% \\
Released random & 16,250 & 15,777 & 97.09\% \\
Official MassSpecGym & 17,556 & 2,956 & 16.84\% \\
\bottomrule
\end{tabular}
}
\end{table}

\textbf{Updated nearest-neighbor baseline with fallback search.}
\citet{khoo2026nearest} subsequently adds a fallback search over the full training library when no same-formula match exists, and introduces a subformula method that combines fingerprints from multiple neighbors. While every test spectrum now receives a prediction, \citet{khoo2026nearest} still reports results on the custom splits with the train-test overlap described above. The subformula method uses the known precursor formula to annotate peaks, averages neighboring fingerprints, and introduces learnable parameters that are tuned on the validation set. We reproduce the \texttt{subformula\_only} variant,\footnote{Implementation of the new nearest neighbor: \href{https://github.com/serenaklm/ML_MS_analysis/tree/6817e9bd994f6915d09b3bb540a9e95e52fb27b4/benchmarked_models/nearest_neighbour}{\texttt{ML\_MS\_analysis} at commit \texttt{6817e9b}}.} evaluating it across all queries in the official MassSpecGym~v1.5 retrieval setting. Under this leakage-free split, the updated nearest-neighbor baseline remains substantially outperformed by machine learning methods, including MIST \textit{4M-aug}, ICEBERG, and GLACIER.

\textbf{Features, neighbor selection, and fingerprint prediction.}
For each spectrum, we retain the most intense peaks (at most 100) and assign fragment subformulas of the precursor ion within a mass tolerance of $\max(20\,\mathrm{ppm}, 0.01\,\mathrm{Da})$. Fragment-formula and neutral-loss-formula tokens are weighted by square-root intensity, taking the maximum weight for repeated tokens, and $L_2$-normalized. Query tokens absent from the training vocabulary are discarded before normalization. Training spectra are ranked by token-vector cosine similarity plus $\alpha$ times the heteroatom-weighted Bray--Curtis similarity of molecular-formula element counts, with weight $w$ for heteroatoms and 1 for carbon and hydrogen. We average the radius-2, 4096-bit Morgan fingerprints (without chirality) of the top-$k$ training spectra using softmax weights with temperature $T$. The required hyperparameter tuning is performed on all 19,429 official validation spectra by mean fingerprint Jaccard similarity after thresholding at 0.5, using the same grid as the updated implementation: $w\in\{1,2,4,8,16\}$, $\alpha\in\{0,0.25,0.5,1,2,4\}$, $T\in\{0.01,0.02,0.05,0.1,0.2,0.5\}$, and $k\in\{1,3,5,10,20,50\}$. This gives $w=2$, $\alpha=2$, $T=0.1$, and $k=50$.

\textbf{Retrieval evaluation.}
For retrieval, we retain the continuous weighted-average fingerprint and rank the official formula-based candidates by Tanimoto similarity, preserving the supplied candidate order for ties. All 17,556 test spectra are evaluated, including those without a same-formula training spectrum. Hit rates at 1, 5, and 20 are 13.13\%, 26.56\%, and 44.30\%, respectively, with MCES@1 of 12.45, as reported in Table~\ref{tab:retrieval_leaderboard_full}. We compute 99.9\% BCa confidence intervals from 20,000 bootstrap resamples of test spectra. Our reproduction is available in \url{https://github.com/harrylaucngd/msms-models}.

\subsection{Detailed Formulations of Artifact-Exploiting Stress-Test Baselines}
\label{sec:appendix_jailbreak}

We provide complete technical formulations for the five artifact-exploiting or spectrum-blind baselines evaluated in Section~\ref{sec:task-validity}, based on deep audits of the corresponding codebases.

\textbf{Method 1: DreaMS + ChemBERTa Alignment.}
This architecture bridges the spectral and molecular modalities via a lightweight trainable adapter between two pretrained encoders. The spectral encoder is DreaMS~\citep{bushuiev2025dreams}, a 7-layer Graphormer-style Transformer with 8 attention heads that encodes MS/MS peaks as ($m/z$, intensity) pairs with Fourier positional features, producing a pooled embedding $\mathbf{z}_s \in \mathbb{R}^{1024}$. The molecular encoder is ChemBERTa-2~\citep{ahmad2022chemberta2} (\texttt{Derify/ChemBERTa\_augmented\_pubchem\_13m}), a RoBERTa-style masked language model pretrained on 13M SMILES strings, whose CLS-token output $\mathbf{z}_m \in \mathbb{R}^{768}$ provides a SMILES-surface representation.

The adapter $f_\phi$, a multi-block MLP with $N=8$ residual blocks (hidden dim 2048), projects the DreaMS embedding into ChemBERTa's latent space: $\hat{\mathbf{z}}_m = f_\phi(\mathbf{z}_s)$. Training jointly optimizes a bidirectional InfoNCE objective with a learnable temperature $\tau$:
\begin{equation}
    \mathcal{L}_{\mathrm{InfoNCE}} = -\frac{1}{2B}\sum_{i=1}^{B}\left[\log\frac{e^{\hat{\mathbf{z}}_{m,i} \cdot \mathbf{z}_{m,i}/\tau}}{\sum_j e^{\hat{\mathbf{z}}_{m,i} \cdot \mathbf{z}_{m,j}/\tau}} + \log\frac{e^{\mathbf{z}_{m,i} \cdot \hat{\mathbf{z}}_{m,i}/\tau}}{\sum_j e^{\mathbf{z}_{m,i} \cdot \hat{\mathbf{z}}_{m,j}/\tau}}\right],
\end{equation}
where the sum is over all $B$ items in a batch, and the positive pair $(i, i)$ is the spectrum--ground-truth-SMILES pair. A Supervised Contrastive (SupCon) loss with isomer-aware negatives is optionally added to handle structural ambiguity. Only the adapter $f_\phi$ and the last two DreaMS Transformer layers are trained; ChemBERTa remains completely frozen. At retrieval, candidate molecules are embedded by ChemBERTa and ranked by cosine similarity to the adapter output $\hat{\mathbf{z}}_m$.

Critical failure mechanism: because ChemBERTa's training is sensitive to SMILES token-level syntax, and the ground-truth SMILES consistently originates from non-RDKit sources (retaining PubChem-style conventions), the adapter learns to produce $\hat{\mathbf{z}}_m$ that matches the \emph{syntactic formatting} of the positive pair, rather than encoding mass spectral information.

\textbf{Method 2: Chemformer Two-Stage Retrieval.}
This two-stage pipeline builds upon a frozen Chemformer BART encoder and a jointly trained spectral encoder.

\textit{Stage 1 (Pre-Retrieval):} The molecular encoder is a frozen Chemformer encoder (BART architecture: $d=512$, 6 layers, 8 heads, vocabulary size 523 SMILES tokens), producing a CLS embedding $\mathbf{e}_m = f_{\mathrm{mol}}(\mathbf{x}_{\mathrm{SMILES}}) \in \mathbb{R}^{512}$. The spectral encoder is an FC$(2 \to 512)$ projection followed by a 6-layer Transformer encoder with mean pooling: $\mathbf{e}_s = f_{\mathrm{spec}}(\mathbf{x}_{\mathrm{peaks}}) \in \mathbb{R}^{512}$. Both are $L_2$-normalized before computing similarities. Training optimizes a dual-path InfoNCE loss at $\tau = 0.1$:
\begin{equation}
    \mathcal{L}_{\mathrm{mol2ms}} = \mathrm{CE}\big([\mathbf{e}_{m,i} \cdot \mathbf{e}_{s,i},\; \mathbf{e}_{m,i} \cdot \mathbf{e}_{s,i}^-] / \tau,\; 0\big),\quad \mathcal{L}_{\mathrm{ms2mol}} = \mathrm{CE}(\mathbf{E}_s \mathbf{E}_m^T / \tau,\; \mathbf{I}),
\end{equation}
where $\mathbf{e}_{s,i}^-$ is an augmented negative spectrum for molecule $i$, and $\mathcal{L}_{\mathrm{ms2mol}}$ is a full-batch CLIP-style contrastive loss. The total loss is $\mathcal{L} = 0.5(\mathcal{L}_{\mathrm{mol2ms}} + \mathcal{L}_{\mathrm{ms2mol}})$.

\textit{Stage 2 (Generative Retrieval):} After Stage 1, the top-$K=40$ candidate molecules are precomputed per spectrum. A cross-fusion module performs cross-attention where the spectral encoder sequence (queries) attends to the concatenated full-sequence hidden states of the $K$ candidates (keys/values), producing a context-enriched spectral representation. A Chemformer decoder (6-layer BART, initialized from pretrained weights) then autoregressively generates a SMILES string conditioned on this cross-fused context via teacher-forcing cross-entropy loss. At inference, beam search (beam size 5) generates a refined candidate, which is then used to re-rank the initial top-$K$ pool by cosine similarity between the generated molecule embedding and each candidate embedding.

Failure propagation across stages: Stage 1 inflates the pre-retrieved candidate pool by exploiting the canonicalization artifact in the InfoNCE training signal. The Stage 2 decoder, conditioned on these biased top-$K$ candidates, further consolidates the shortcut, since the cross-fusion module learns to match the formatting signature of the positive candidate that consistently appears near the top of the Stage 1 ranking.



\textbf{Method 3: ChemBERTa SMILES Classifier.}
A spectrum-blind binary classifier built on top of a pretrained SMILES Transformer~\citep{ahmad2022chemberta2} (\texttt{DeepChem/ChemBERTa-77M-MLM}, loaded via the Hugging Face \texttt{transformers} library with a randomly initialized 2-class classification head) that maps a single SMILES string to the probability that it is a ground-truth query molecule rather than a decoy candidate (i.e., a randomly sampled SMILES from the corresponding candidate list, as per \texttt{MassSpecGym\_retrieval\_candidates\_formula.json}). The model is fine-tuned for two epochs with a binary cross-entropy objective on balanced positive/negative pairs. At test time, the classifier scores each candidate independently and ranks the candidate set in descending order of predicted probability. No spectral input and no per-query ground-truth information are used at inference time.

\textbf{Method 4: PubChem Default Ranking.}
This method exploits the popularity bias described in Section~\ref{sec:task-validity}. For a given query, candidates are ranked in the default order returned by the PubChem PUG-REST API. No spectral information is used. Ground-truth test molecules, being predominantly well-characterized natural products and common metabolites, systematically rank higher in this default order, achieving $>49\%$ hit rate@1. This baseline aims to expose a critical limitation that any benchmark whose ground-truth molecules are non-uniformly sampled from chemical space will be susceptible to this ranking shortcut.

\textbf{Method 5: Ranking by Number of Chiral Atoms.}
A spectrum-blind ranker that exploits a systematic offset between ground-truth molecules and their candidates in the count of annotated stereocenters. For each candidate molecule $c$, the ranker computes
\begin{equation}
n(c) = \sum_{a \in c} \mathbb{1}\bigl\{\texttt{a.GetChiralTag()} \neq
\texttt{Chem.ChiralType.CHI\_UNSPECIFIED}\bigr\},
\end{equation}
the number of atoms $a$ in $c$ carrying an explicit stereochemistry annotation in the SMILES (i.e., @ or @@), evaluated via RDKit's \texttt{Chem.MolFromSmiles} followed by \texttt{GetAtoms} and \texttt{GetChiralTag}. Candidates are then ranked by $n(c)$. The ranking direction is decided based on the training set: descending if ground-truth molecules carry on average more annotated stereocenters than their candidates, ascending otherwise. The ranker uses no spectral input and no per-query ground-truth information at test time.

\subsection{Pipeline of Annotating Missing Collision Energies in the MassSpecGym Test Set}
\label{sec:appendix_collsion_energy}

Because collision energy annotations are incomplete in MassSpecGym, but forward models such as ICEBERG and GLACIER require collision energy labels, we use a staged workflow for each model to evaluate it on the full benchmark. We first construct a subset containing spectra with observed collision energies and train the corresponding simulator on its training split: the fragmentation and intensity models for ICEBERG, and the joint fragmentation--intensity model for GLACIER. These models are then used to impute missing collision energies in the original full labels. For each spectrum lacking a collision energy, the corresponding simulator generates predictions over a grid of normalized collision energies from 0\% to 150\% NCE in steps of 5\%, converted to eV using the precursor $m/z$. The collision energy whose predicted spectrum has the highest spectral entropy similarity to the experimental spectrum, excluding the precursor region, is selected as a pseudo-label. The imputed labels are written back into a derived dataset, with retrieval candidate tables updated to use the same collision-energy values. Finally, each simulator is retrained on the completed training split and evaluated on the full test split. The ICEBERG pipeline includes fragmentation training, intensity training, and contrastive intensity fine-tuning; GLACIER uses joint fragmentation--intensity training without contrastive fine-tuning. This two-pass design preserves calibration from experimentally annotated spectra while allowing both models to be evaluated on the full MassSpecGym test set. These ICEBERG and GLACIER results in Table~\ref{tab:retrieval_leaderboard_full} should not be confused with the ``simulation\_challenge'' setting, where only a subset of training and testing data is involved.

\subsection{Formal Metric Definitions for Reproducible Evaluation}
\label{sec:appendix_metrics}

We provide precise, implementation-pinned definitions for each evaluation metric referenced in Section~\ref{sec:infrastructure}.

\textbf{InChIKey Hit Rate.}
An InChIKey is a 27-character identifier whose first 14 characters encode molecular connectivity. Since MS/MS fragmentation is largely insensitive to stereocenters, two structures sharing the same connectivity layer should be considered equivalent. Formally, a prediction $\hat{m}$ is a hit for ground truth $m^*$ if $\text{InChIKey}_{14}(\hat{m}) = \text{InChIKey}_{14}(m^*)$, where $\text{InChIKey}_{14}(\cdot)$ denotes the first 14 characters of the InChIKey. Using the full 27-character InChIKey incorrectly rejects stereoisomers as non-hits and---in data splitting---fails to exclude structurally similar training molecules that differ only in stereochemistry.

\textbf{MCES Distance.}
Given two molecular graphs $G_1=(V_1,E_1)$ and $G_2=(V_2,E_2)$, the Maximum Common Edge Subgraph (MCES) is the largest subgraph (by edge count) isomorphic to a subgraph of both. The MCES distance is:
\begin{equation}
    d_\text{MCES}(G_1, G_2) = |E_1| + |E_2| - 2\,|\text{MCES}(G_1, G_2)|.
\end{equation}
Since computing MCES exactly is NP-hard, we use \href{https://github.com/AlBi-HHU/myopic-mces}{\texttt{myopic-mces}}~1.0.0 with PuLP~2.9.0 and the \texttt{PULP\_CBC\_CMD} solver. We set \texttt{threshold=15} and \texttt{always\_stronger\_bound=True}, with a solver time limit of 5 seconds. Stereochemistry is ignored during graph comparison.

\textbf{Tanimoto Fingerprint Similarity.}
Let $A$ and $B$ be binary fingerprint vectors of a predicted and ground-truth molecule. Tanimoto similarity is:
\begin{equation}
    T(A, B) = \frac{\sum_i A_i B_i}{\sum_i A_i + \sum_i B_i - \sum_i A_i B_i}.
\end{equation}
Fingerprints must be ECFP4 Morgan fingerprints computed with radius $r=2$ and $n=2048$ bits (via \texttt{rdkit.Chem.AllChem.GetMorganFingerprintAsBitVect} or \texttt{rdkit.Chem.rdFingerprintGenerator.FPType.MorganFP}).

\textbf{Close and Meaningful Match Rates.}
The Close Match and Meaningful Match metrics are complementary structural similarity metrics introduced outside the original MassSpecGym benchmark in \citep{Butler2023ms2mol}. A \textit{Close Match} requires $T(A, B) \geq 0.675$, and a \textit{Meaningful Match} requires $T(A, B) \geq 0.40$, where $T$ is Tanimoto similarity computed with the default \textbf{RDKit} fingerprint instead of ECFP4 (i.e., via \texttt{rdkit.Chem.rdFingerprintGenerator.FPType.RDKitFP}). Any deviation from the RDKit fingerprint should be explicitly reported.

\textbf{Confidence Intervals.}
Per-example metric values (e.g., per-molecule InChIKey hits or per-prediction Tanimoto similarities) are aggregated by their mean across the test set. To quantify uncertainty around this mean, BCa bootstrap confidence intervals are computed: the test-set values are resampled with replacement 20{,}000 times, the mean is recomputed on each resample, and bias-corrected and accelerated percentile levels are used to obtain the lower and upper bounds of the 99.9\% confidence interval. We use \texttt{scipy.stats.bootstrap} with \texttt{confidence\_level=0.999}, \texttt{n\_resamples=20000}, \texttt{method='BCa'}, and a fixed random seed for reproducibility.

\subsection{Metric-Implementation Sensitivity Tables}
\label{sec:appendix_metric_gaps}

Table~\ref{tab:metric_variants} shows the influence of evaluation metric parameters on FRIGID performance. High deviation in the results highlights the importance of keeping metric implementations consistent with the original reference implementations (e.g., the ones implemented in the MassSpecGym codebase).

\begin{table*}[htbp!]
\centering
{\caption{\textbf{Comparison of Metric Variants on Identical FRIGID Predictions for the MassSpecGym \textit{de novo} Generation Challenge w/ \textit{Bonus}.} Predictions are from FRIGID with MIST \textit{4M-aug} after 10 refinement rounds on all 17{,}556 test spectra; parentheses show 99.9\% BCa confidence intervals (20{,}000 spectrum-level resamples, seed 42). ``r'' denotes the Morgan fingerprint radius, ``thld'' denotes the MCES \texttt{threshold} parameter below which exact computation is attempted, and ``bound'' denotes the \texttt{always\_stronger\_bound} parameter in the MCES implementation. Black rows show the reference implementation; gray rows show alternative implementations.}
\label{tab:metric_variants}}
{
{\small
\begin{tabular}{llcc}
\toprule
Metric & Variant & Top-1 & Top-10 \\
\midrule
Accuracy & InChIKey-14 & 9.05 (8.35--9.74)\% & 12.23 (11.42--13.04)\% \\
\color{gray} Accuracy & \color{gray} Full InChIKey & \color{gray} 4.97 (4.45--5.54)\% & \color{gray} 6.80 (6.15--7.47)\% \\
\color{gray} Tanimoto & \color{gray} Morgan $r=2$, 1024 bits & \color{gray} 0.3969 (0.3906--0.4033) & \color{gray} 0.4435 (0.4371--0.4501) \\
Tanimoto & Morgan $r=2$, 2048 bits & 0.3873 (0.3808--0.3938) & 0.4328 (0.4262--0.4394) \\
\color{gray} Tanimoto & \color{gray} Morgan $r=2$, 4096 bits & \color{gray} 0.3827 (0.3763--0.3893) & \color{gray} 0.4276 (0.4210--0.4344) \\
\color{gray} Tanimoto & \color{gray} Morgan $r=3$, 1024 bits & \color{gray} 0.3399 (0.3336--0.3463) & \color{gray} 0.3871 (0.3805--0.3938) \\
\color{gray} Tanimoto & \color{gray} Morgan $r=3$, 2048 bits & \color{gray} 0.3257 (0.3193--0.3322) & \color{gray} 0.3712 (0.3645--0.3780) \\
\color{gray} Tanimoto & \color{gray} Morgan $r=3$, 4096 bits & \color{gray} 0.3189 (0.3124--0.3255) & \color{gray} 0.3635 (0.3566--0.3705) \\
\color{gray} MCES & \color{gray} thld=100, bound=False & \color{gray} 37.99 (37.03--38.89) & \color{gray} 28.90 (28.05--29.72) \\
\color{gray} MCES & \color{gray} thld=15, bound=False & \color{gray} 12.56 (12.40--12.72) & \color{gray} 11.32 (11.17--11.47) \\
MCES & thld=15, bound=True & 12.60 (12.45--12.76) & 11.41 (11.26--11.56) \\
\bottomrule
\end{tabular}%
}
}
\end{table*}

\subsection{Agentic Infrastructure for Automated Leaderboard Submission and Code Review}\label{sec:appendix_code-review}

\begin{figure}[t!]
\centering
\includegraphics[width=1\linewidth]{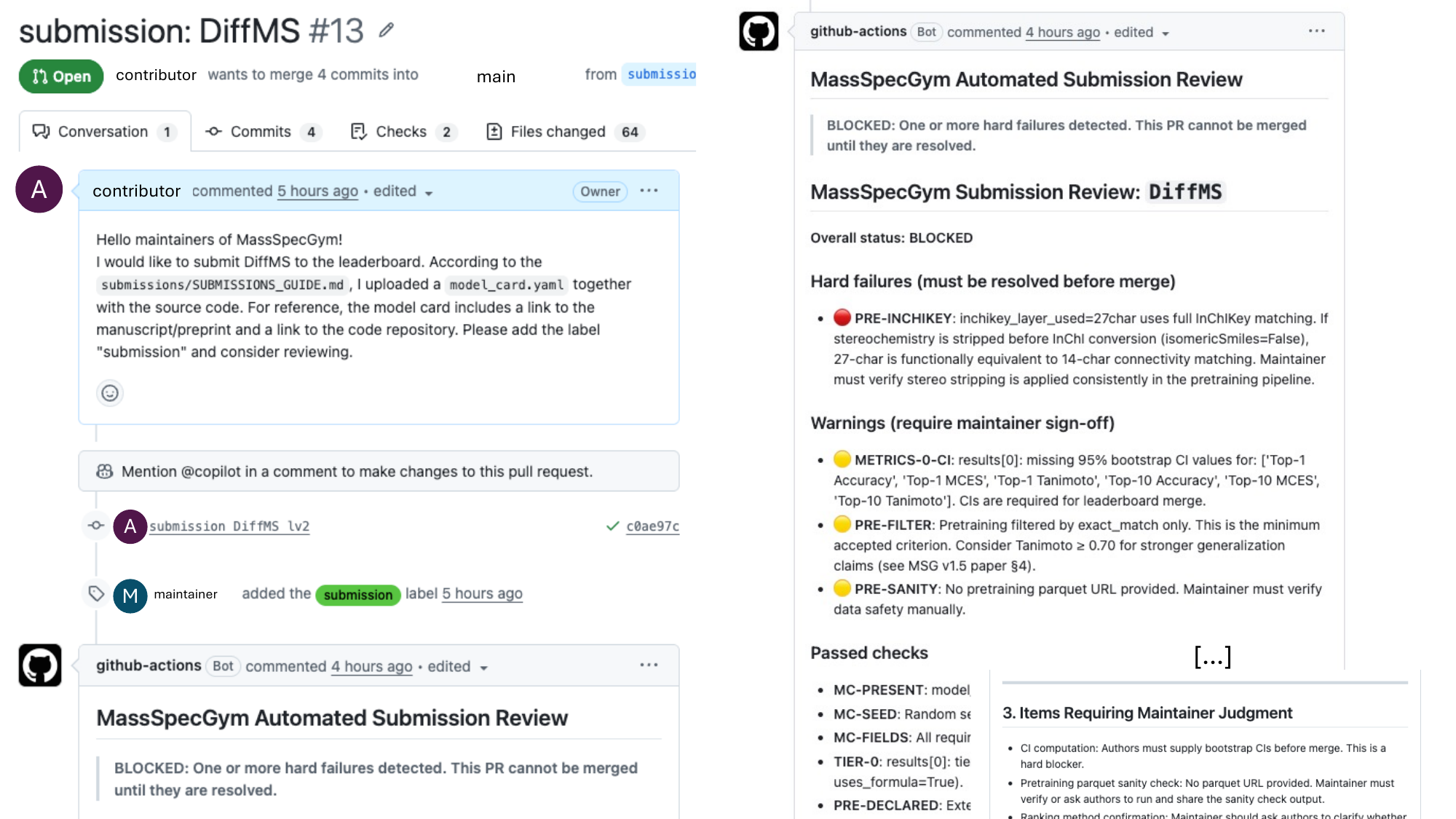}
\caption{Example 1 of an LLM-enabled workflow for reviewing submissions to the MassSpecGym leaderboard. When a contributor submits a pull request for their model to be featured in the leaderboard, an automated GitHub Action workflow is triggered that parses through the provided model card and source code to review according to the guidelines outlined in the provided skill (\textit{vide supra}) based on this publication. The output of the workflow is a set of errors, warnings and recommendations for a human maintainer. Once approved, the source code files are deleted, the model card is kept and the leaderboard updated.}
\label{fig:code_review_diffms}
\end{figure}

\begin{figure}[t!]
\centering
\includegraphics[width=1\linewidth]{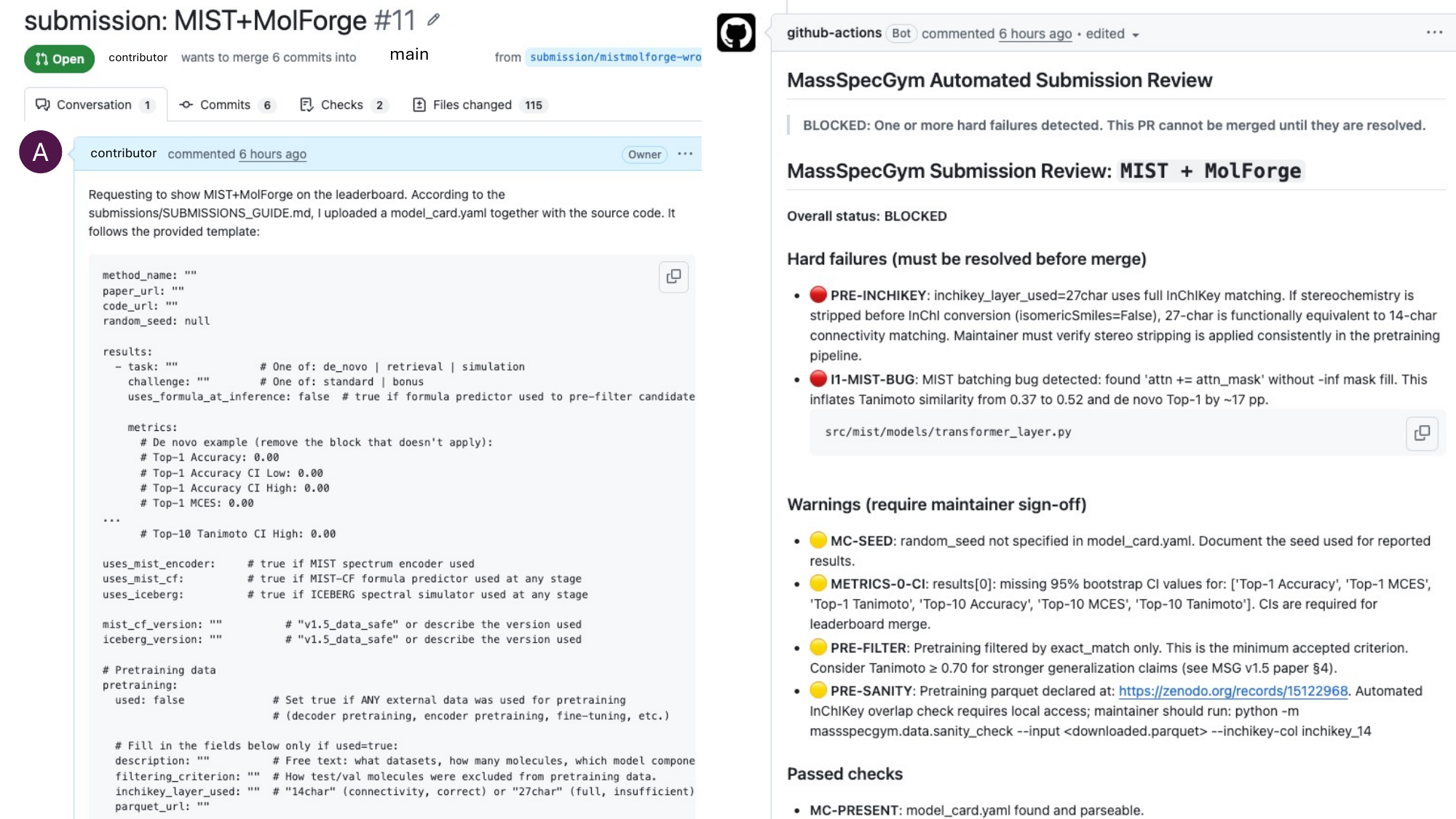}
\caption{Example 2 of an LLM-enabled workflow for reviewing submissions to the MassSpecGym leaderboard. Similarly to Figure~\ref{fig:code_review_diffms}, the output of the workflow is a set of errors, warnings and recommendations for a human maintainer. The workflow correctly identifies data leakage due to stereoisomers and the MIST inference bug.}
\label{fig:code_review_mistmolforge}
\end{figure}

To operationalize the recommendations of this audit and reduce the burden of manual review, MassSpecGym~v1.5 introduces a structured submission workflow for new leaderboard entries. Authors submit a pull request to the main repository containing the relevant results, code, and a model card that specifies method metadata including paper and code URLs, pretraining data and filtering criteria, oracle component versions, and evaluation setup. On every pull request, a GitHub Actions workflow automatically runs a review script, which performs an analysis and heuristic checks covering all failure modes identified in this paper and generates a narrative review of the method. Hard failures must be resolved unless maintainers document an override; warnings require explicit maintainer sign-off. A second workflow runs on every merge to \texttt{main} and validates the full set of results for schema consistency and confidence interval coverage. A maintainer review guide documents residual judgment calls beyond a static analysis, which includes reading the paper methods section for implicit formula use, verifying pretraining data safety against the InChIKey exclusion list, and spot-checking reproducibility. The system prompt and skill to guide the code review are shown below. Examples of identifying issues in a submission are shown in Figures~\ref{fig:code_review_diffms} and~\ref{fig:code_review_mistmolforge}. 
This resource is intended to support both the maintainers and users of MassSpecGym with updating the leaderboard and contributors with additional guardrails and feedback in the development of new models.

\begin{tcolorbox}[enhanced,breakable,mytitle,myprompt,title=SKILL.md,top=0.6em,bottom=0.4em,left=0.6em,right=0.6em]
\begin{Verbatim}[breaklines,breakanywhere,fontsize=\small,commandchars=!\{\}]
---
name: review
description: Maintainer guide for reviewing MassSpecGym leaderboard submissions. Covers interpreting the automated review report, performing judgment calls the automation cannot make, and approving or rejecting PRs. The automated review (scripts/review_submission.py, triggered by the review_submission GH Action) handles deterministic checks; this guide covers the residual human review.
---

# MassSpecGym Submission Review — Maintainer Guide

## Role of this guide

Every leaderboard PR triggers `scripts/review_submission.py` automatically and posts a structured report as a PR comment. That report handles all deterministic checks (schema, CIs, tier integrity, MIST bug, pretraining filter, metric overrides). **Your job as maintainer is to:**

1. Read the automated report and triage any WARNINGs that require judgment
2. Fetch and read the paper if the automated LLM review flagged concerns or couldn't access the paper
3. Check items the automation explicitly cannot verify (listed below)
4. Approve or request changes

Hard failures in the automated report must be resolved by the author before you even look at the PR. Do not override hard failures without a documented reason.

---

## Submission requirements (what a valid PR must contain)

1. A new row in the correct `results/*.csv` with all required metrics and 99.9% BCa bootstrap CIs
2. A `submissions/<method_name>/model_card.yaml` filled from the template

The method name in the CSV `Method` column must exactly match `method_name` in `model_card.yaml` (underscored folder name, spaced method name). See `submissions/SUBMISSION_GUIDE.md`.

---

## Step 1: Read the automated report

The PR comment from the review bot contains:

- **Hard failures** — must be fixed; CI blocks merge
- **Warnings** — require your sign-off; use a PR review comment to document your decision
- **LLM narrative review** — treat as a second opinion; verify any specific issues it flags

For warnings, document your reasoning inline on the PR before approving.

---

## Step 2: Things the automation cannot check — do these manually

### 2a. Paper methods section vs. model card

Read the paper's methods/data section and cross-check against `model_card.yaml`:

- Does the paper describe using ground-truth test formulas, directly or to filter candidates, in the **mass-based** (standard) challenge? If yes and the model is submitted to `results/retrieval.csv` (not !textit{bonus}), reject.
- Does the paper describe pretraining on data sources *not listed* in the model card? Flag the discrepancy.
- Does the paper use ICEBERG-generated spectra for pretraining? Check which ICEBERG version (data-safe v1.5 or upstream). If unclear, ask the author.
- Does the paper's reported number match the CSV entry? Values that differ from the paper by >0.5 pp on the primary metric need explanation.

### 2b. Spectrum-blind shortcut check (S2)

If the model is a retrieval model, mentally check: could a model that ignores the spectrum entirely (using the PubChem PUG-REST API's default return order or SMILES format) achieve similar results?

The PubChem default-ranking baseline achieves 49.98% hit rate@1 on the formula-based retrieval task. If the submission's hit rate@1 is at or below this level, it may not be learning from spectra at all. Ask the author to compare against the PubChem default-ranking baseline if not already included.

**Spectrum-blind classifier test (run if suspicious):** The ChemBERTa SMILES classifier achieves 99.66% hit rate@1 on non-canonicalized candidate sets. If the candidate set for this submission is non-standard, ask the author to confirm the v1.5 pre-canonicalized candidate set is used.

If you have access to compute: train or run a version of the model with the spectral input zeroed out or permuted. If performance stays substantially above random (~1/pool_size), the model likely exploits a spurious correlation and should be investigated further before acceptance.

### 2c. Pretraining data — if no parquet provided

If `pretraining.used=true` but no parquet URL is given, ask the author to either:
- Provide a public parquet for the InChIKey overlap check, or
- Run `python -m massspecgym.data.sanity_check --input their_data.parquet --inchikey-col inchikey_14` themselves and share the output

Do not accept claims of data safety without this check if pretraining is declared.

### 2d. Reproducibility spot-check

If the submission includes a checkpoint or inference script, spot-check one metric:

```bash
# In the massspecgym conda environment (Python 3.11)
git clone <author-fork-url>
cd MassSpecGym
pip install -e ".[dev]"
python scripts/run.py \
    --job_key review_run \
    --run_name <model_name>_review \
    --test_only \
    --no_wandb \
    --seed 42
```

A discrepancy >0.5 pp on the primary metric requires explanation. If the author cannot provide a reproducible checkpoint, note this in your review but it is not grounds for automatic rejection — reproducibility is aspirational for submissions without public checkpoints.

---

## Step 3: Metric specifications (reference)

All submitted metrics must match these pinned implementations:

| Metric | Specification |
|--------|--------------|
| InChIKey hit rate | First 14 characters (connectivity layer) only — **not** full 27-char |
| Tanimoto similarity | Morgan ECFP4, radius=2, 2048 bits |
| MCES distance | `threshold=15`, `always_stronger_bound=True` — see `massspecgym/utils.py:MyopicMCES` |
| Cosine similarity | Standard MS/MS cosine as in `massspecgym` |
| Jensen-Shannon similarity | As in `massspecgym` |

**Do not accept** metric implementations reimported from third-party codebases unless independently verified against the specifications above.

---

## Step 4: Required metrics per task

### De novo (standard and !textit{Bonus})
`Top-1 Accuracy`, `Top-1 MCES`, `Top-1 Tanimoto`, `Top-10 Accuracy`, `Top-10 MCES`, `Top-10 Tanimoto` — all required, all with CIs.

### Retrieval (standard)
`Hit rate @ 1`, `Hit rate @ 5`, `Hit rate @ 20`, `MCES @ 1` — all required, all with CIs.

### Retrieval (!textit{Bonus})
`Hit rate @ 1`, `Hit rate @ 5`, `Hit rate @ 20`, `MCES @ 1` — all required, all with CIs.

### Simulation (standard and !textit{Bonus})
`Cosine Similarity`, `Jensen-Shannon Similarity`, `Hit rate @ 1`, `Hit rate @ 5`, `Hit rate @ 20` — all required, all with CIs.

Standard and !textit{bonus} variants must not be compared in the same table row. A model submitted to both must have two separate CSV rows.

---

## Step 5: MIST batching bug (I1) — background

If a submission uses the MIST encoder from an external codebase (not `massspecgym/models/encoders/mist/`), verify the attention masking in batched inference. The original MIST encoder added `attn += attn_mask` without first converting the boolean mask to `-inf`, which allows padding tokens to contribute to the softmax. The v1.5 MassSpecGym MIST encoder has this fixed.

**Observed checkpoint issue:** For MIST vDiffMS + MolForge, increasing inference batch size from 1 to 24 raises encoder Tanimoto similarity from 0.457 to 0.812 and de novo top-1 accuracy from 10.73% to 31.75%. The documented training recipe does not reproduce this inflation.

**What to look for:**
```python
# BUG (wrong):
attn += attn_mask

# SUBTLE BUG — fix code present but result discarded:
new_attn_mask = torch.zeros_like(attn_mask, dtype=q.dtype)
new_attn_mask.masked_fill_(attn_mask, float("-inf"))
attn += attn_mask   # still uses raw bool mask, new_attn_mask silently ignored

# FIX (correct):
new_attn_mask = torch.zeros_like(attn_mask, dtype=q.dtype)
new_attn_mask.masked_fill_(attn_mask, float("-inf"))
attn += new_attn_mask
```

The v1.5 MassSpecGym MIST encoder (`massspecgym/models/encoders/mist/transformer_layer.py`) already applies the fix. If the submission imports from there, this is satisfied.

---

## Step 6: Data leakage — gradient of stringency

Data leakage is not binary. Performance can remain sensitive to filtering out close structural analogs even after exact test matches are excluded. This means a model may claim "clean" data (exact-match excluded) while still benefiting substantially from near-neighbor contamination.

**What to report in your review:**
- What filtering criterion is declared in the model card?
- Is this consistent with what the paper states?
- Does the paper compare against baselines at the same filtering level?

The leaderboard accepts exact-match exclusion as the minimum. If the criterion is weaker or unstated, request clarification before approval.

---

## Final approval checklist

Before approving the PR, confirm:

- [ ] Automated report has no hard failures (or failures have been resolved and re-reviewed, or explicitly overridden with a documented reason)
- [ ] All warnings have been signed off with a documented reason
- [ ] Paper methods section consistent with model card
- [ ] No ground-truth test formula use in a mass-based submission (from reading the paper)
- [ ] Pretraining data safety confirmed (parquet sanity check run or author confirmed)
- [ ] If MIST encoder used: v1.5 fix confirmed or external repo verified
- [ ] If MIST-CF or ICEBERG used: v1.5 data-safe version confirmed
- [ ] All required metrics present with CIs
- [ ] Method name in CSV matches model card exactly
- [ ] `df_test_path` configured so per-sample predictions are saved (recommended)
- [ ] Seed documented
\end{Verbatim}
\end{tcolorbox}

\newpage

\end{document}